\pdfoutput=1
\documentclass[11pt]{article}

\usepackage{latex/acl}

\usepackage{times}
\usepackage{latexsym}

\usepackage[T1]{fontenc}
\usepackage[utf8]{inputenc}

\usepackage{microtype}

\usepackage{graphicx}
\usepackage{import}
\usepackage{inconsolata}

\usepackage{kotex}
\usepackage{amsmath}
\usepackage{mathtools}
\usepackage{amssymb}
\usepackage{amsthm}
\usepackage{multirow}
\usepackage{makecell}
\usepackage{svg}
\usepackage{subfigure}
\usepackage{subcaption}
\usepackage{verbatim}
\usepackage{longtable}
\usepackage{pifont}
\newcommand{\cmark}{\ding{51}}%
\newcommand{\xmark}{\ding{55}}%

\title{RSCF: Relation-Semantics Consistent Filter for Entity Embedding 
 of Knowledge Graph}

\author{Junsik Kim, Jinwook Park, 
Kangil Kim\thanks{\; Corresponding author.}
\\
AI Graduate School \\
Gwangju Institute of Science and Technology \\
  \texttt{junsikkim@gm.gist.ac.kr},
  \texttt{jinwookpark@gm.gist.ac.kr}, \\
  \texttt{kangil.kim.01@gmail.com} \\}

\begin{document}
\maketitle
\begin{abstract}

In knowledge graph embedding, leveraging relation specific entity transformation has markedly enhanced performance. However, the consistency of embedding differences before and after transformation remains unaddressed, risking the loss of valuable inductive bias inherent in the embeddings. 
This inconsistency stems from two problems. First, transformation representations are specified for relations in a disconnected manner, allowing dissimilar transformations and corresponding entity embeddings for similar relations.
Second, a generalized plug-in approach as a SFBR (Semantic Filter Based on Relations) disrupts this consistency through excessive concentration of entity embeddings under entity-based regularization, generating indistinguishable score distributions among relations. 
In this paper,
we introduce a plug-in KGE method, \textit{Relation-Semantics Consistent Filter} (RSCF). 
Its entity transformation has three features for enhancing semantic consistency:
1)  shared affine transformation of relation embeddings across all relations, 2)  rooted entity transformation that adds an entity embedding to its change represented by the transformed vector, and 3) normalization of the change to prevent scale reduction.
To amplify the advantages of consistency that preserve semantics on embeddings, RSCF adds relation transformation and prediction modules for enhancing the semantics. 
In knowledge graph completion tasks with distance-based and tensor decomposition models, 
RSCF significantly outperforms state-of-the-art KGE methods, showing robustness across all relations and their frequencies.
\end{abstract}

\section{Introduction}

Knowledge graphs (KGs) play crucial roles in a wide area of machine learning and its applications~\cite{ zhang2022kcd, zhou2022eventbert,  geng2022path}. However, KGs, even on a large scale, still suffer from incompleteness~\cite{dong2014knowledge}. This problem has been extensively studied as a task to predict missing entities, known as  knowledge graph completion (KGC).

An effective approach for KGC is knowledge graph embedding (KGE) that learns vectors to represent entities and relations in a low dimensional space to measure the validity of triples.
Two primary approaches to determine the validity are distance-based model (DBM) using the Minkowski distance and tensor decomposition model (TDM) regarding KGC as a tensor completion problem~\cite{zhang2020duality}. 

A recently tackled issue of the models is to learn only single embedding for an entity, which is insufficient to express its various attributes in complex relation patterns such as 1-N, N-1 and N-N~\cite{chao2021pairre, ge2023compounding}.
A proposed and effective approach for this issue is entity transformation based model (ETM) that uses relation specific transformations to generate different entity embeddings for relations from their original embedding, enabling more complex entity and relation learning~\cite{ge2023compounding}.

ETMs, however, have a limit to learning useful inductive bias  that could be obtained in semantically similar relations. 
For example, SFBR, a recently proposed method plugged in to various KGE models~\cite{liang2021semantic}, assigns mutually disconnected relation specific transformation to each relation.
Furthermore, under a significantly useful regularizer such as DURA~\cite{zhang2020duality}, especially on TDM, the method critically concentrates entity embeddings, including unobserved entities and generates indistinguishable score distributions across relations. 
Both issues are interpreted as limited learning an important and implicit inductive bias that semantically similar relation have similar relation specific entity transformation, called \textit{relation-semantics consistency} in this paper. 

To alleviate the issues, we present a simple and effective method, \textit{Relation-Semantically Consistent Filter} (RSCF). 
Its entity transformation has three features for enhancing semantic consistency. 1) shared affine transformation for consistency mapping of relations to entity transformations, 2) rooted entity transformation using the affine transformation to generate only the change of an entity embedding subsequently added by this embedding and 3) normalization of the change for preventing critical scale reduction breaking consistency.
To amplify the benefit of the consistency, RSCFs adds relation transformation (RT) and relation prediction (RP)~\cite{chen2021relation}, for inducing useful relation specific semantics on embeddings.

Our contributions are as follows.
\begin{itemize}
    \item We raise and clarify two problems in terms of \textit{relation semantics consistency} in learning useful inductive bias on embeddings.
    \item  We propose a novel and significantly outperforming RSCF as a plug-in KGE method, which induces the consistency and effectively learns useful semantic representations.
    \item We provide experimental results on common benchmarks of KGC, and in-depth analysis to verify the causes and derived effects. 
\end{itemize}

\section{Loss of Useful Inductive Bias} 
Because semantically similar relations have similar embedding~\cite{zhang2018knowledge}, we define that mapping relation embeddings to entity transformations (ETs) is \textit{relation semantically consistent} if and only if any relation pairs $(r_1, r_2)$ and shorter pair $(r_1, r_3)$ for a given $r_1$ are mapped to ET pair $(T_1, T_2)$ and shorter pair $(T_1, T_3)$, respectively. This consistency serves as an inductive bias implying that semantically similar relations have similar ETs and, therefore, overall similar entity embeddings (EEs). Two phenomena of losing this inductive bias and their causes are as follows. 

\paragraph{Disconnection of Entity Transformations}
Disconnected ET loosely use this bias, especially under lack of triplet data. 
In existing methods, relation specific ETs use separate parameters such as $\mathsf{h_r}= \mathsf{W}_{r}\mathsf{h}$ and $\mathsf{t_r}= \mathsf{W}_{r}\mathsf{t}$, where $\mathsf{h}, \mathsf{t}$, are head and tail entity embedding, and $\mathsf{W_{r}}$ is a relation specific transformation. 
Despite the disconnection, the methods can still learn similar $\mathsf{W_{r}}$ for given two similar relation embeddings if their desirable entity ranks are similar.
However, limited observation of entities due to sparse KG introduces a wide variety of possible ETs and their corresponding embedding distributions, thereby diluting consistency.
In this environment, the disconnected representation without any specific training and initialization process aiming to foster the consistency is exposed to the loss of useful inductive bias of similar relations. 

\begin{table}[t]
\centering
\tiny
\begin{tabular}{lcccc}
\hline

Metric                 & ET-SFBR & ET-RSCF & EE-SFBR & EE-RSCF \\ \hline
Intra Cluster Distance ($\downarrow$)   & 1.10    & 0.47    & 2.35    & 0.53    \\ 
Inter Cluster Distance ($\uparrow$) & 0.27    & 0.82     & 0.40    & 0.85    \\ \hline
\end{tabular}
\caption{Intra  and Inter Cluster Distance of ET and EE. RSCF shows more concentrated yet distinguishable clusters compared to SFBR. (Intra Cluster Distance: mean distance of elements in a cluster to their centroid, Inter Cluster Distance: distance of a centroid with its closest centroid. Both measures are averaged across clusters.)}

\label{tab:transformation_head_SC_ICD}
\end{table}

\begin{figure}[t]
\centering
\includegraphics[width=0.5\textwidth]{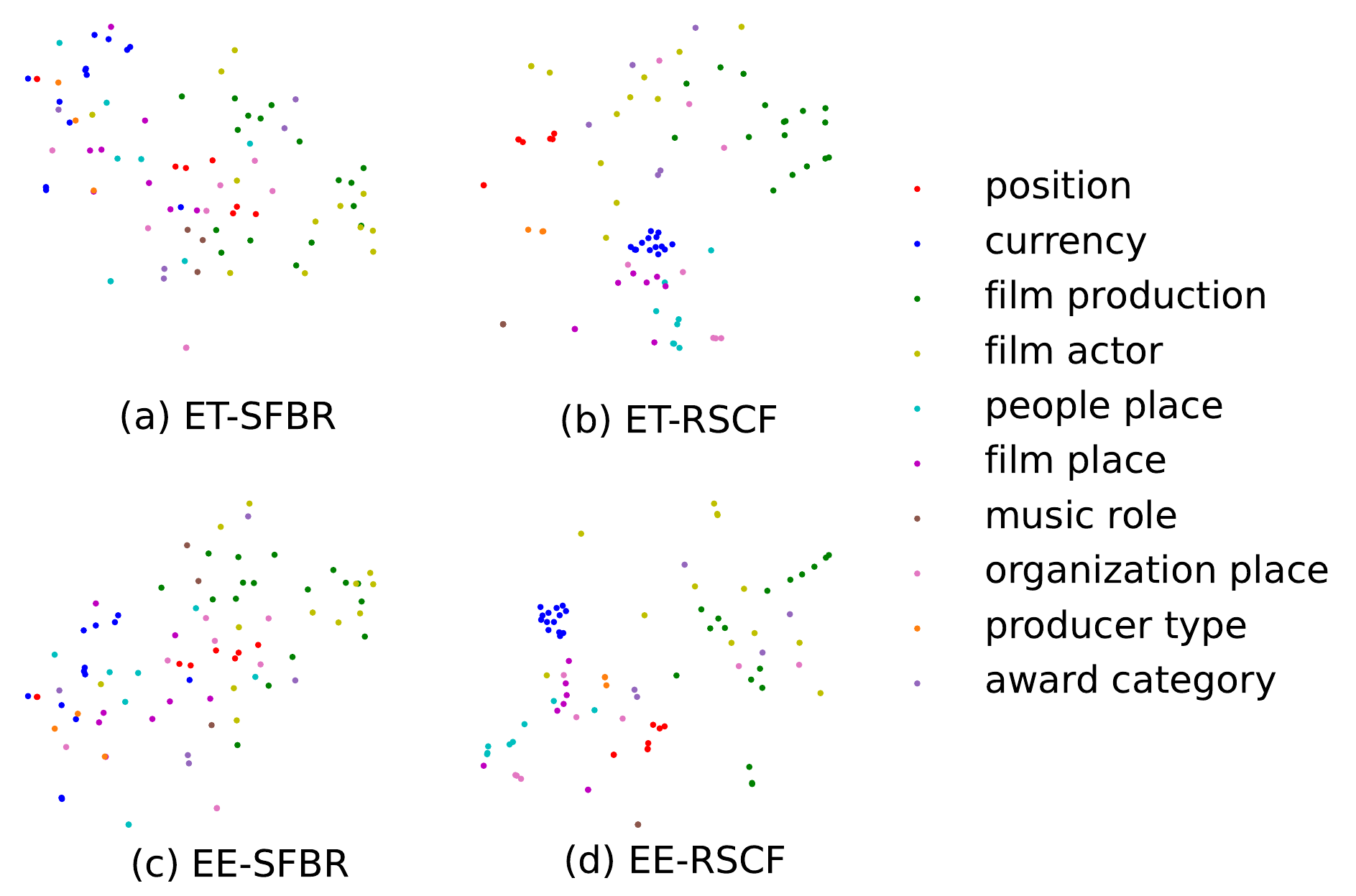}
\caption{t-SNE result distribution of head ET ((a) and (b)) and EE ((c) and (d)) for semantically similar relation groups. Same color represents same semantic group.}
\label{fig:transformation-head}
\end{figure}

\paragraph{Empirical Evidence for Disconnection}
The quantitative empirical evidence of disconnection in ET and EE is shown in Table~\ref{tab:transformation_head_SC_ICD}.
The results represent that SFBR applied to TransE produces ET and EE that are less distinguishable among clusters (lower Inter Cluster Distance) and less concentrated in each cluster (higher Intra Cluster Distance) compared to RSCF. Additionally, Figure~\ref{fig:transformation-head} shows a qualitative example of the distributions visualized by t-SNE. This example also indicates the evidence of disconnection of ET and EE in SFBR.
More details of making groups for ET and EE, Intra Cluster Distance, and Inter Cluster Distance are shown in Appendix~~\ref{appendix: A}.

\begin{figure}
    \subfigure[Score Distribution]{
        \includegraphics[width=0.8\columnwidth]{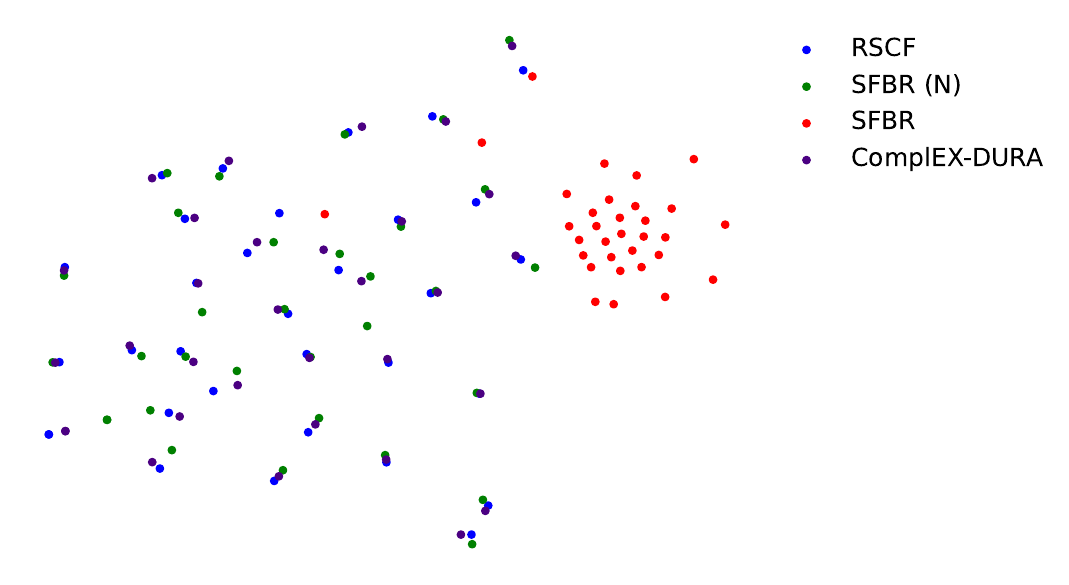}
    }
    \hfill
    \subfigure[MRR of validation set (left) and Transformation Scale (right)]{
        \includegraphics[width=0.85\columnwidth]{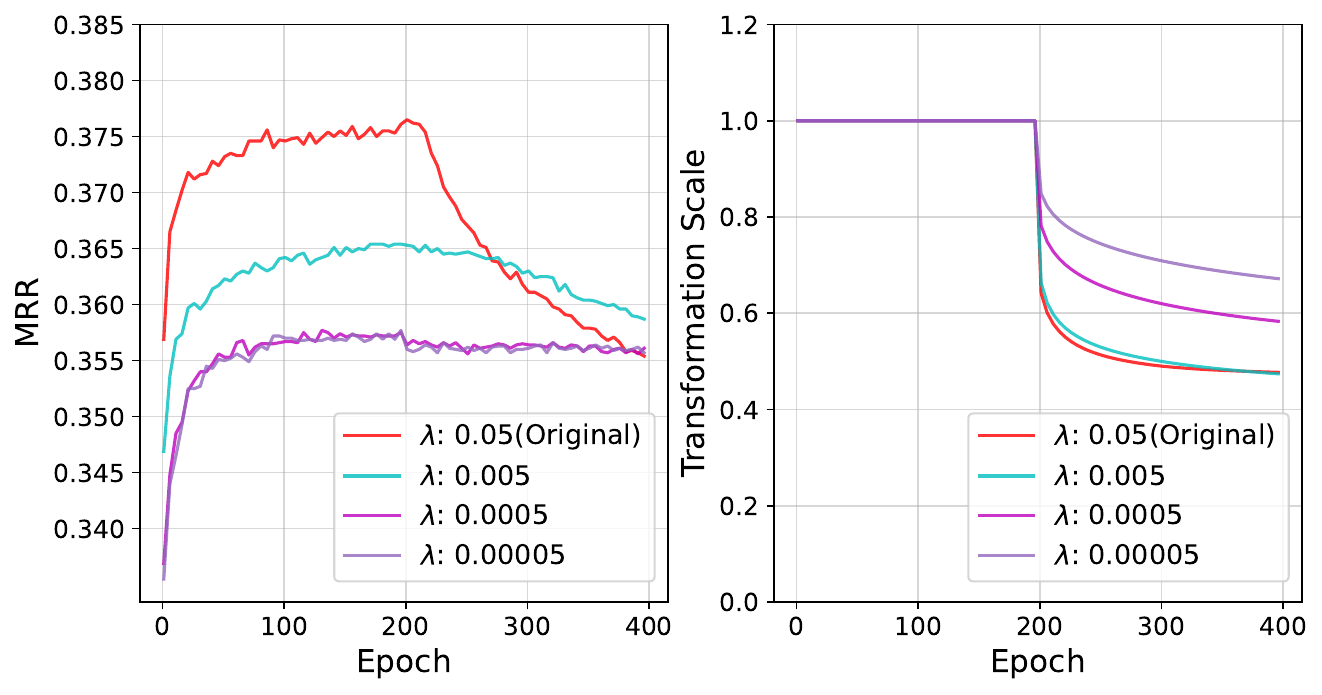}
    }
    \centering
    \caption{Result of entity embedding concentration, and performance and scale decrease in training. The results are collected from ComplEX with DURA regularization. DURA is applied in all epochs and SFBR is applied after 200 epochs ($\lambda$:  regularization weight).}
    \label{fig:Score Over Smoothing}
\end{figure}
    
\begin{figure*}[ht]
    \centering
    \includegraphics[width=0.9\textwidth]{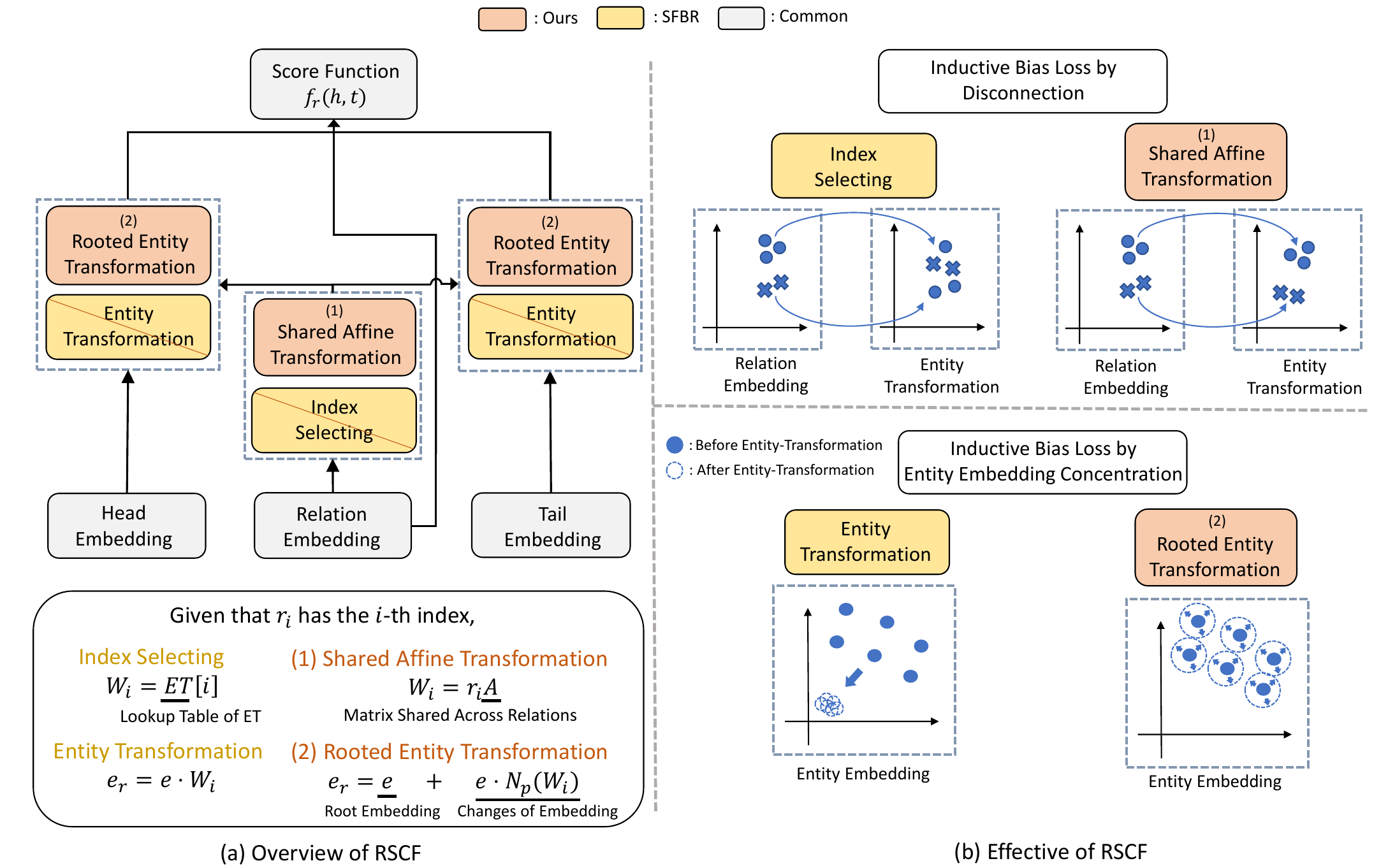}
    \caption{Overview of RSCF and its effect. Its process (left) is illustrated on SFBR coloring changed modules. The two effects (right) are shown by comparing SFBR and RSCF on ET and entity embeddings.}
    \label{fig:method_overview}
\end{figure*}

\paragraph{Entity Embedding Concentration}
In particular, SFBR additionally loses consistency under entity-based regularization, DURA~\cite{zhang2020duality}. 
In KGE based on TDM, DURA has shown significant improvement enough to be inevitable. However, ComplEX-SFBR with DURA reduces the scale of ET, causing a strong concentration of entire entity embeddings. Observed entities are relatively safe because the score distribution is continuously adjusted to predict correct triples, but unobserved entities are critically vulnerable to the concentration causing indistinguishable score distributions for semantically different relations, implying critically broken consistency. 
This cause of this phenomenon is simply derived in the following equations of DURA in the original (above)~\cite{zhang2020duality} and DURA in SFBR (below).

\begin{equation}
\begin{array}{lr}
{\scriptstyle \sum_{p} } & {\scriptstyle ||\mathsf{h_i \overline{R_j}}||^2_2 + ||\mathsf{h_i}||^2_2  + || \mathsf{t_k}||^2_2 
    + ||\mathsf{t_k \overline{R_j}^{\top}}||^2_2}  \\
{\scriptstyle \sum_{p}}&{\scriptstyle ||\mathsf{W_{r_j}h_i \overline{R_j}}||^2_2  + ||\mathsf{W}_{r_j}\mathsf{h_i}||^2_2 + || \mathsf{t_k}||^2_2 + ||\mathsf{t_k \overline{R_j}^{\top}}||^2_2}

\end{array}
\label{eq:DURA_regularizer}
\end{equation}
where $p = (h_i, r_j, t_k) \in S$ for  total training data $S$, $\mathbf{h_i}$ and $\mathbf{t_k}$ are head and tail embeddings with indices and $\mathbf{\overline{R_j}}$ is a matrix representing relation $r_j$. 
In the equation~\ref{eq:DURA_regularizer}, to minimize DURA loss, model always decreases the scale of ET (simple proof in Appendix~\ref{appendix: Proof of scale decrease of ET}) %~\ref{appendix:scale decrease of ET}) 
and this causes indistinguishable score distribution in all score distributions.
\paragraph{Empirical Evidence for Concentration}
Figure~\ref{fig:Score Over Smoothing} presents a T-SNE visualization of score distributions for selected queries. We selected the relation $r_1$, which shows significantly low performance in SFBR on FB15k-237, and selected all queries ($h$, $r_1$, ?) for this relation $r_1$ in the validation set. We then generate score distribution for each query using ComplEX-RSCF, ComplEX-SFBR, ComplEX-SFBR with normalization (SFBR (N)), and the ComplEX-DURA. The results show that SFBR concentrates embeddings into a small cluster, while the other methods are diversely dispersed. 
\paragraph{Do We Need to Use DURA regularizer?}
Generating indistinguishable score distributions cannot be merely resolved by handling the regularization weight. Figure~\ref{fig:Score Over Smoothing} (b) shows the valid MRR (left) of SFBR and transformation scale (right) according to the regularizer weight $\lambda$. 
In training until 200 epochs, largely weighted DURA shows significant performance, but applying SFBR starts to decrease MRR and the transformation scale. 
The results imply that integrating SFBR with DURA causes performance degradation with scale decrease ending up in the entity embedding concentration. 
Also, the result of SFBR with a small weighted DURA indicates that simply excluding DURA on the TDM will critically decrease the performance. 

\section{Method}

\paragraph{Overview}
In this section, we propose \textit{Relation-Semantics Consistent Filter} (RSCF) to address the consistency issues. 
In Figure~\ref{fig:method_overview}, the overall filtering process of RSCF, distinguished features compared to ETMs, and their intended effects are illustrated.

%\subsection{Entity Transformation}
RSCF represents the ET as an addition of original embedding (\textcircled{c}) and its relation specific change. The change is generated by an affine transformation from relation embedding (\textcircled{a}), and then normalized (\textcircled{b}), described as 
\begin{align}
    \begin{split}
    \mathbf{e_r} &= 
    ( \overbrace{\mathrm{N}_p}^{\text{\textcircled{b}}}(\overbrace{\mathbf{rA_1}}^{\text{\textcircled{a}}})  \overbrace{+ \mathbf{1}}^{\text{\textcircled{c}}}) \otimes \mathbf{e} \\
    \end{split}
    \label{eq:main_method_final}
\end{align}
where $\mathbf{A_1} \in \mathbf{R}^{n\times n}$ is shared affine transformation across all relations, $\mathbf{r}$ and $\mathbf{e} \in \mathbf{R}^{n}$are relation and entity embedding. $\mathrm{N}_p(\mathbf{rA}) = \frac{\mathbf{rA}}{\lVert \mathbf{rA}\rVert_{p}}$, and $\otimes$ is an elementwise product.
Detailed motivation and effects are as follows.

\paragraph{Shared Affine Transformation for Consistency}

A basic property of affine transformation is to maintain the parallelism of two parallel line segments after the transformation and preserves the ratio of their lengths. 
This property guarantees consistent mapping of relation embeddings at least on a line to generated vectors (part \textcircled{a} in Equation~\ref{eq:main_method_final}). 
Even in the case that embeddings are not exactly on a line, the consistency is maintained with high probability as shown in Table~\ref{tab:Monte-carlo simulation}. It presents the proportion of consistency maintenance rate for each component of RSCF based on Monte Carlo simulations. The result indicates that our ET can preserve consistency in the most cases. (Details about this result are presented in Appendix~\ref{appendix: Monte Carlo}).

After normalization of the generated vectors (part \textcircled{b}), the consistency still holds in most cases, showing a maximum of over 99\% in the table~\ref{tab:Monte-carlo simulation}.
The addition of one vector to the normalized change (part \textcircled{c}) does not alter the inequality of distances, so the consistency is again maintained. 
Overall, by applying the affine transformation, we can maintain the consistency between relation embedding and its ET. To implement the affine transformation shared across relations, we simply adopt a linear transformation for $\mathbf{A}$.

\begin{table}[t]
\centering
\resizebox{\columnwidth}{!}{%
\begin{tabular}{lllccc}
\hline
Method                    & On a Line                          & $\frac{|\overline{AC}|}{|\overline{AB}|} > 1$ & $\frac{|\overline{AC}|}{|\overline{AB}|} > 1.01$ &   $\frac{|\overline{AC}|}{|\overline{AB}|} > 1.02$ & \\ \hline
Transformation (\textcircled{a}) & \multicolumn{1}{c}{1.000} & \multicolumn{1}{c}{.728} & \multicolumn{1}{c}{.808} & \multicolumn{1}{c}{.875} \\

Normalization (\textcircled{b}) & \multicolumn{1}{c}{$.965$} & \multicolumn{1}{c}{$.869$} &\multicolumn{1}{c}{$.958$} & \multicolumn{1}{c}{$.994$} \\

Add one (\textcircled{c}) & \multicolumn{1}{c}{1.000} & \multicolumn{1}{c}{1.000} & \multicolumn{1}{c}{1.000} & \multicolumn{1}{c}{1.000}\\
\hline

\end{tabular}%
}

\caption{
Consistency of our ET: The numbers represent the success rates of preserving the superiority of distances for 10,000 randomly generated samples of three relation embeddings, notated as A, B, and C. (Line: the rates for samples with elements exactly on a line, the others: the rate for samples not on a line with varying distance rate conditions.)
}
\label{tab:Monte-carlo simulation}
\end{table}

\paragraph{Rooted Entity Transformation}
Sharing an affine transformation across all relations inevitably reduces the expressiveness of ET compared to entirely separate relation specific ET such as SFBR. To mitigate the negative effects from this reduction, we decrease required expressiveness by learning only the changes in entity embeddings, rather than learning their diverse positions.
Moreover, this rooted ET representation enables safely bounding changes via normalization without altering original entity embeddings. \footnote{Refer to Appendix~\ref{appendix: Normalization of Change for Reducing Entity Embedding Concentration} for details on the bounds of entity changes.}
To implement it, we add one to the normalized change $\mathbf{N}_p\mathbf{(rA)}$  and multiply it to the original entity embedding (part \textcircled{c}).

\begin{table*}
\centering
\scriptsize
\begin{tabular}{lccclcc}
\hline
\multirow{2}{*}{Model} & \multicolumn{3}{c}{ET Features}                        & \multirow{2}{*}{Entity Transformation} & \multirow{2}{*}{Relation Transformation} & \multirow{2}{*}{Relation Prediction~\cite{chen2021relation}} \\ \cline{2-4}
                       & \multicolumn{1}{c}{\textcircled{a}}               & \multicolumn{1}{c}{\textcircled{b}}              & \multicolumn{1}{c}{\textcircled{c}}               &                                        &                                          &                                     \\ \hline
PairRE                 & \xmark                 & \xmark                 & \xmark                & $\mathbf{e_r} = \mathbf{e} \otimes r^e $  & \xmark 
& \xmark

\\ 
SFBR                   & \xmark                 & \xmark                 & \xmark                 & $\mathbf{e_r} = \mathbf{W_r} \cdot \mathbf{e} + \mathbf{b}$  & \xmark 
&
\xmark
\\ 
CompoundE              & \xmark                   & \xmark                  & \xmark                   & $\mathbf{e_r} = \mathbf{T_r} \cdot \mathbf{R_r}(\theta) \cdot  \mathbf{S_r} \cdot  \mathbf{e}$  & \xmark 
& \xmark

\\  \hline
RSCF                   & \cmark                 & \cmark                 & \cmark                 & $\mathbf{e_r} = \psi(r) \otimes \mathbf{e}$  & $\mathbf{r_{ht}} = \psi'(h) \otimes \psi'(t) \otimes \mathbf{r}$ 
& \cmark
                                  \\ \hline
                      & \multirow{2}{*}{}

\end{tabular}
\caption{
Summary of difference between RSCF and ETMs ($\mathbf{h}$: head entity, $\mathbf{t}$: tail entity, $\mathbf{e_r}$: transformed entity from $\mathbf{h}$ and $\mathbf{t}$, which contains both transformed head entity $\mathbf{h_r}$ and transformed tail entity $\mathbf{t_r}$, 
$\psi$: ET represented in Equation~(\ref{eq:main_method_final}),
$\psi'$: applied ET to relation transformation shown in  and Equation~(\ref{eq:main_method_relation_final}), 
\textcircled{a}: shared affine transformation, 
\textcircled{b}: bounding change from $\mathbf{e}$, 
\textcircled{c}: rooting $\mathbf{e_r}$ to $\mathbf{e}$). Computational complexity is presented in Appendix~\ref{appendix: Complexity Analsis}}
\label{tab:RSCF_SFBR_comparison}
\end{table*}

\paragraph{Relation Prediction 
for More Consistent Relation Embedding to its Semantics} 
%Enhanced Relation Semantics on Relation Embedding
%to Facilitate Relation Semantic Reflection to Relation Embeddings}
The inductive bias introduced by the RSCF is dependent on the semantics of relation embeddings. Therefore, directly enhancing these semantics results in the improvement of RSCF performance. An effective approach is Relation Prediction (RP)~\cite{ chen2021relation} forming a cluster for semantically similar relations and improving discrimination of dissimilar relations. We add the training objective of RP~\cite{chen2021relation} to RSCF as follows:

\begin{small}
    \begin{align}
    \begin{split}
    \mathcal{L} = \sum_p  \phi(\mathbf{h_r} | \mathbf{r_{ht}}, \mathbf{t_r}) + \phi(\mathbf{t_r} | \mathbf{h_r}, \mathbf{r_{ht}}) + \lambda\phi(\mathbf{r} | \mathbf{h}, \mathbf{t}) 
    \end{split}
    \label{eq:trainig_objective_of_RSCF}
    \end{align}
    \label{eq:RP_equation}
\end{small}
where $\phi$ is a loss function with a score function and $\lambda$ is a hyper-parameter that controls the contribution of RP.

\paragraph{Relation Transformation 
for Relation Embedding of its Fine-Grained Semantics}

In KGs, some relations have various semantic meanings that can be divided into fine-grained sub-relations according to their semantics~\cite{zhang2018knowledge}. 
Because the semantic meanings of sub-relations are determined by their context, which is defined by head and tail entities~\cite{jain2022discovering}, 
we propose an entity specific relation transformation (RT) to split relations into sub-relations, and apply the filter of ET of RSCF for the same purpose. By using Equation~\ref{eq:main_method_final}, we present the RT as follows:

{\small
    \begin{align}
        \begin{split}
        \mathbf{r_{ht}} &= (\mathrm{N}_p(\mathbf{hA_2}) + 1) \otimes (\mathrm{N}_p(\mathbf{tA_3}) + 1) \otimes \mathbf{r}
        \end{split}
        \label{eq:main_method_relation_final}
    \end{align}
}

where $\mathbf{A_2} \in \mathbf{R}^{n\times n}$ and $\mathbf{A_3} \in \mathbf{R}^{n\times n}$ are shared affine transformation across all heads and tails. To predict score of given triplet $(h, r, t)$, transformed entities $\mathbf{e_r}$ and relation $\mathbf{r_{hr}}$ are used. The difference of RSCF and ETMs and are summarized in Table~\ref{tab:RSCF_SFBR_comparison}.

\section{Related Works}
\paragraph{Knowledge Graph Embedding} 
KGE encodes entities and relations into low-dimensional latent spaces to assess the validity of triples. TransE~\cite{bordes2013translating} and RotatE~\cite{sun2018rotate} describe each relation as a translation and rotation between entities, respectively. DistMult~\cite{yang2015embedding} regards KGC as a tensor completion problem in euclidean space, and ComplEX~\cite{trouillon2016complex} extends it to complex space. TransHRS~\cite{zhang2018knowledge} improves knowledge representation by using the information from the HRS. DURA~\cite{zhang2020duality}, AnKGE~\cite{yao2023analogical}, and CompliE~\cite{cui2024modeling} are methods that can be applied to KGE models to prevent overfitting, provide analogical inference and enable composition reasoning. VLP~\cite{li2023copy} presents an explicit copy strategy to allow referring to related factual triples. GreenKGC~\cite{wang2023greenkgc} and SpeedE~\cite{pavlovic2024speede} propose low-dimensional embedding methods to handle large-scale KGs. WeightE~\cite{zhang2023weighted} utilizes a reweighting technique to alleviate the data imbalance issue. UniGE~\cite{liu2024bridging} introduce integration KGE in both euclidean and hyperbolic to capture various relational patterns. However, using only a single embedding for an entity or a relation can restrict the learning of complex relation patterns.

\begin{table*}[t]
\centering
\scriptsize
\begin{tabular}{lccccccccc}
\hline
\multicolumn{1}{c}{\multirow{2}{*}{\textbf{Knowledge Graph Embedding}}} & \multicolumn{3}{c}{\textbf{WN18RR}} & \multicolumn{3}{c}{\textbf{FB15k-237}}        & \multicolumn{3}{c}{\textbf{YAGO3-10}}         \\ \cline{2-10} 
\multicolumn{1}{c}{}                                                    & MRR        & H@1        & H@10      & MRR           & H@1           & H@10          & MRR           & H@1           & H@10          \\ \hline

DistMult~\cite{yang2015embedding}                                                                & .430       & .390       & .490      & .241          & .155          & .419          &     -          &     -          &    -           \\
ComplEX~\cite{trouillon2016complex}                                                                 & .440       & .410       & .510      & .247          & .158          & .428          &     -          &   -            & -              \\
RotatE~\cite{sun2018rotate}                                                                  & .476       & .428       & .571      & .338          & .241          & .533          &         -      &      -         &    -           \\
DistMult-HRS~\cite{zhang2018knowledge}                                                                  & -       & -       & -      & .315          & .241          & .496          &         -      &      -         &    -           \\
AutoETER~\cite{niu2020autoeter}                                                                  & -       & -       & -      & .344          & .250          & .538          &         .550      &      .465         &    .699           \\

PairRE~\cite{chao2021pairre}                                                                  &    -        &      -      &     -      & .351          & .256          & .544          &      -         &       -        & -              \\

CIBLE~\cite{cui2022instance}                                                                   & .490       & .446       & .575      & .341          & .246          & .532          &      -     &     -      &     -      \\
ReflectE~\cite{zhang2022knowledge}                                                                   & .488       & .450       & .559      & .358          & .263          & .546          &      -     &    -       &     -      \\
HAKE-AnKGE~\cite{yao2023analogical}                                                                    & .500       & .454      & .587   & \underline{.385}         & \underline{.288}         & \underline{.572}        & -          & -          & -          \\
CompoundE~\cite{ge2023compounding}                                                               & .491       & .450       & .576      & .357          & .264          & .545          & -    & -    & -     \\
RotatE-GreenKGC~\cite{wang2023greenkgc}                                                         & .411       & .367       & .491      & .345          & .265          & .507          & .453          & .361          & .629          \\
RotatE-VLP~\cite{li2023copy}                                                              & .498       & \underline{.455}       & .582      & .362          & .271          & .542          &  -             &     -          &    -           \\
RotatE-WeightE~\cite{zhang2023weighted}                                                              & .501       & .448       & \textbf{.592}      & .371          & .281          & .557          &  .580             &     .504          &    .713           \\
CompliE-DURA~\cite{cui2024modeling}                                                            & .495       & .453       & .579      & .372          & .277          & .563          &      -         &   -            &     -          \\
SpeedE~\cite{pavlovic2024speede}                                                                  & .493       & .446       &      -     & .320          & .227          &      -         & .413          & .332          &     -          \\
UniGE~\cite{liu2024bridging}                                                                   & \underline{.502}       & \underline{.455}       & \textbf{.592}      & .357          & .264          & .559          & .583          & .512          & \underline{.715}          \\
\hline

TransE~\cite{bordes2013translating}                                                               & .226       & -           & .501      & .294          &   -            & .465          &   -            &      -         & -              \\

ComplEX-DURA~\cite{zhang2020duality}                                                          & .491       & .449       & .571      & .371          & .276          & .560          & \underline{.584}          & .511          & .713          \\

\hline

TransE-SFBR~\cite{liang2021semantic}                                                      & .242       & .028       & .548      & .338        & .240        & .538       &     -        &      -      &     -       \\

ComplEX-DURA-SFBR~\cite{liang2021semantic}                                                                   & .498       & .454       & .584      & .374          & .277          & .567          & \underline{.584}          & \underline{.512}          & .712          \\  \hline
{\multirow{2}{*}{TransE-RSCF (Ours)}}                                                            & .267   & .066       & .546      & .363 & .264 & .558 & - & - &  - \\
& 	$\pm$.001       & 	$\pm$.002       & 	$\pm$.002      & 	$\pm$.001 & 	$\pm$.001 & 	$\pm$.001 & - & - &  - \\

{\multirow{2}{*}{ComplEX-DURA-RSCF (Ours)}}                                                            & \textbf{.503}       & \textbf{.460}       & \underline{.588}      & \textbf{.388} & \textbf{.295} & \textbf{.573} & \textbf{.589} & \textbf{.516} & \textbf{.718}  \\ 
    & $\pm$.001       & $\pm$.001       & $\pm$.002      & $\pm$.001 & $\pm$.002 & $\pm$.004 & $\pm$.002 & $\pm$.003 & $\pm$.002  \\
\hline
\end{tabular}
\caption{Test performance of KGE-based KGC on FB15k-237, WN18RR and YAGO3-10. Bold indicates the best result, and underlined signifies the second best result. $\pm$ indicates standard deviation. (The comparison results of RSCF, SFBR, and other KGE models presented in Appendix~\ref{appendix: Performance Compariosn of RSCF}.)}%~\ref{appendix:Reproduce of SFBR in TDM}).}
\label{tab:KGE-results}
\end{table*}

\paragraph{Entity Transformation Models} ETM is a model that uses relation specific ET to model various attributes of an entity. Models such as TransH~\cite{wang2014knowledge}, TransR~\cite{lin2015learning}, and TransD~\cite{ji2015knowledge} are variants of TransE~\cite{bordes2013translating}, designed to handle complex relations by employing hyperplanes, projection matrices, and dynamic mapping matrices for their transformation functions, respectively. Recently, AutoETER~\cite{niu2020autoeter} learns the type embedding for each entity with relation specific transformation. PairRE~\cite{chao2021pairre} performs a scaling operation through the Hadamard product to the head and tail entities. SFBR~\cite{liang2021semantic} and AT~\cite{yang2021improving} present a universal entity transformation applicable to both DBM and TDM. ReflectE~\cite{zhang2022knowledge} introduces relation specific householder transformation to handle sophisticated relation mapping properties.  CIBLE~\cite{cui2022instance} use relation-aware-transformation for prototype modeling to represent the knowledge graph. CompoundE~\cite{ge2023compounding} applied compound operation to both head and tail entities. However, these models have no chance for inductive bias sharing due to the separate parameter of ET, and SFBR~\cite{liang2021semantic}, which can be applied to both DBM and TDM, suffers from indistinguishable score distribution because of the entity embedding concentrations.

\section{Experiments}

\subsection{Settings}

\paragraph{Dataset} To evaluate our proposed RSCF models, we consider three KG datasets: WN18RR~\cite{dettmers2018convolutional}, FB15k-237~\cite{toutanova2015observed}, and YAGO3-10~\cite{mahdisoltani2013yago3}. The statistics for the three benchmarks are shown in Appendix~\ref{appendix: dataset statistics}.

\paragraph{Evaluation Protocol} 
We evaluated the performance of KGC following the filtered setting~\cite{bordes2013translating}. The filtered setting removes all valid triples from the candidate set when evaluating, except for the predicted triple. We adopt the MRR and Hits@N to compare the performance of different KGE models. MRR is the average of the inverse mean rank of the entities and Hits@N is the proportion of correct entities ranked within top k.

\paragraph{Baselines and Training Protocol}
We compare the performance of RSCF with the KGE models: TransE, DistMult, ComplEX, RotatE,
DistMult-HRS, AutoETER, ComplEX-DURA, PairRE, SFBR, CIBLE, ReflectE, HAKE-AnKGE, CompoundE, RotatE-GreenKGC, RotatE-VLP, RotatE-WeightE, CompliE-DURA, SpeedE, and UniGE. 

Because RSCF is a module that is plugged in based on existing models, we use DBM, including TransE, RotatE, and TDM, including CP, RESCAL, and ComplEX as base models. Additionally, following the setting of SFBR, ET is applied to both head and tail entities in DBM, while it is applied only to the head entity in TDM due to computational cost~\cite{liang2021semantic}. For the same reason, both head and tail entities are utilized for RT in DBM, while only the head entity is used in TDM. In the FB15k-237, the entity/relation ratio and the triple/relation ratio are significantly lower than in the other two datasets, limiting the context information available to each relation. This limitation is particularly critical in TDM, which relies solely on the head entity. Therefore, RT is not applied to TDM in the FB15k-237 dataset.

\subsection{Performance}

%WORKING POINT
\paragraph{Performance on KGC}
Table~\ref{tab:KGE-results} shows the performance comparison of the RSCF and other KGE models on WN18RR, FB15k-237 and YAGO3-10. 
Overall, RSCF shows higher or competitive performance compared to base models like TransE and ComplEX-DURA and other KGE models. Especially in FB15k-237 and YAGO3-10, RSCF outperforms other state-of-the-art models that include HAKE-AnKGE and CompliE-DURA.

\begin{table*}[t]
\resizebox{\textwidth}{!}{
\centering
\begin{tabular}{llc}
\hline
Query (h, r, ?) $\vert$ Correct Answer                                                                          & Related Triples in Training Set                                                           & Rank(R/S)               \\ \hline
\multirow{2}{*}{(Guillermo del Toro, \textit{/people/person/place\_of\_birth}, ?) $\vert$ \textbf{Guadalajara}} & (Guillermo del Toro, \textit{/people/person/places\_lived./people/place\_lived/location}, \textbf{Jalisco}) & \multirow{2}{*}{\textbf{5} / 35} \\    & (Guillermo del Toro, \textit{/people/person/nationality}, \textbf{Mexico})                                  &                         \\
\multirow{2}{*}{(Shawn Pyfrom, \textit{/people/person/places\_lived./people/place\_lived/location}, ?) $\vert$ \textbf{Florida}}  & (Shawn Pyfrom, \textit{/people/person/place\_of\_birth}, \textbf{Tampa})                                    & \multirow{2}{*}{\textbf{3} / 32} \\    & (Shawn Pyfrom, \textit{/people/person/nationality}, \textbf{United States of America})                      &                         \\
\multirow{2}{*}{(Walt Whitman, \textit{/people/person/places\_lived./people/place\_lived/location}, ?) $\vert$ \textbf{New York}} & (Walt Whitman, \textit{/people/deceased\_person/place\_of\_death}, \textbf{Camden})                         & \multirow{2}{*}{\textbf{10} / 21} \\    & (Walt Whitman, \textit{/people/person/nationality}, \textbf{United States of America})                      &                         \\ \hline
\end{tabular}}
\caption{Example KGC results of RSCF compared to SFBR (R: rank of RSCF, S: rank of SFBR). Related triples show that similar relations to the queries have similar entities to the correct answers in the training set. TransE is used as baseline}
\label{tab:positive example}
\end{table*}

\begin{table}[h]
\centering
\tiny
\begin{tabular}{lcccccc}
\hline
\multicolumn{1}{c}{\multirow{2}{*}{\textbf{Model}}} & \multicolumn{2}{c}{\textbf{ET}}               & \multirow{2}{*}{\textbf{RP}} & \multirow{2}{*}{\textbf{RT}} & \textbf{T} & \textbf{C} \\ \cline{2-3} \cline{6-7} 
\multicolumn{1}{c}{}                                & \multicolumn{1}{l}{\textcircled{a}} & \multicolumn{1}{l}{\textcircled{b} \& \textcircled{c}} &                              &                              & MRR                  & MRR                   \\ \hline

RSCF                                                &   \cmark                      &   \cmark                      &  \cmark                    &   \cmark                   & .363                   & .387    
\\ \hline

+ w/o ET                                              &   \xmark                       &  \xmark                        &  \cmark                     &     \cmark                  & .356                                                              & .385                         \\
+ w/o RP                                              &   \cmark                       &  \cmark                        &    \xmark                   &   \cmark                    & .358                                                                & .374                         \\
+ w/o RT                                              &   \cmark                       &  \cmark                        &    \cmark                   & \xmark                      & .356                                      & .388                                   \\
\; + w/o RP                                            &  \cmark                       &    \cmark                     &    \xmark                  &  \xmark                    & .349                                     & .375                                   \\
\; + w/o ET                                            &    \xmark                     &   \xmark                      &    \cmark                  &  \xmark                    & .338                                    & .385                                  \\
\; + w/o \textcircled{a}                                           &   \xmark                      &  \cmark                       &    \cmark                  &   \xmark                   & .354                                     & .386                                        \\
\; + w/o \textcircled{b} \& \textcircled{c}                                         &   \cmark                      &  \xmark                       & \cmark                     &   \xmark                   & .353                    & .377                               \\                  \hline
\end{tabular}
\caption{Results of an ablation study of RSCF on FB15k-237. TransE and ComplEX are used as base models. \textbf{T} and \textbf{C} indicate the base models of RSCF, which denote TransE and ComplEX, respectively. MRR is used for performance comparison.}
\label{tab:ablation}
\end{table}

\begin{figure}[h]
\centering

\includegraphics[width=0.7\columnwidth]{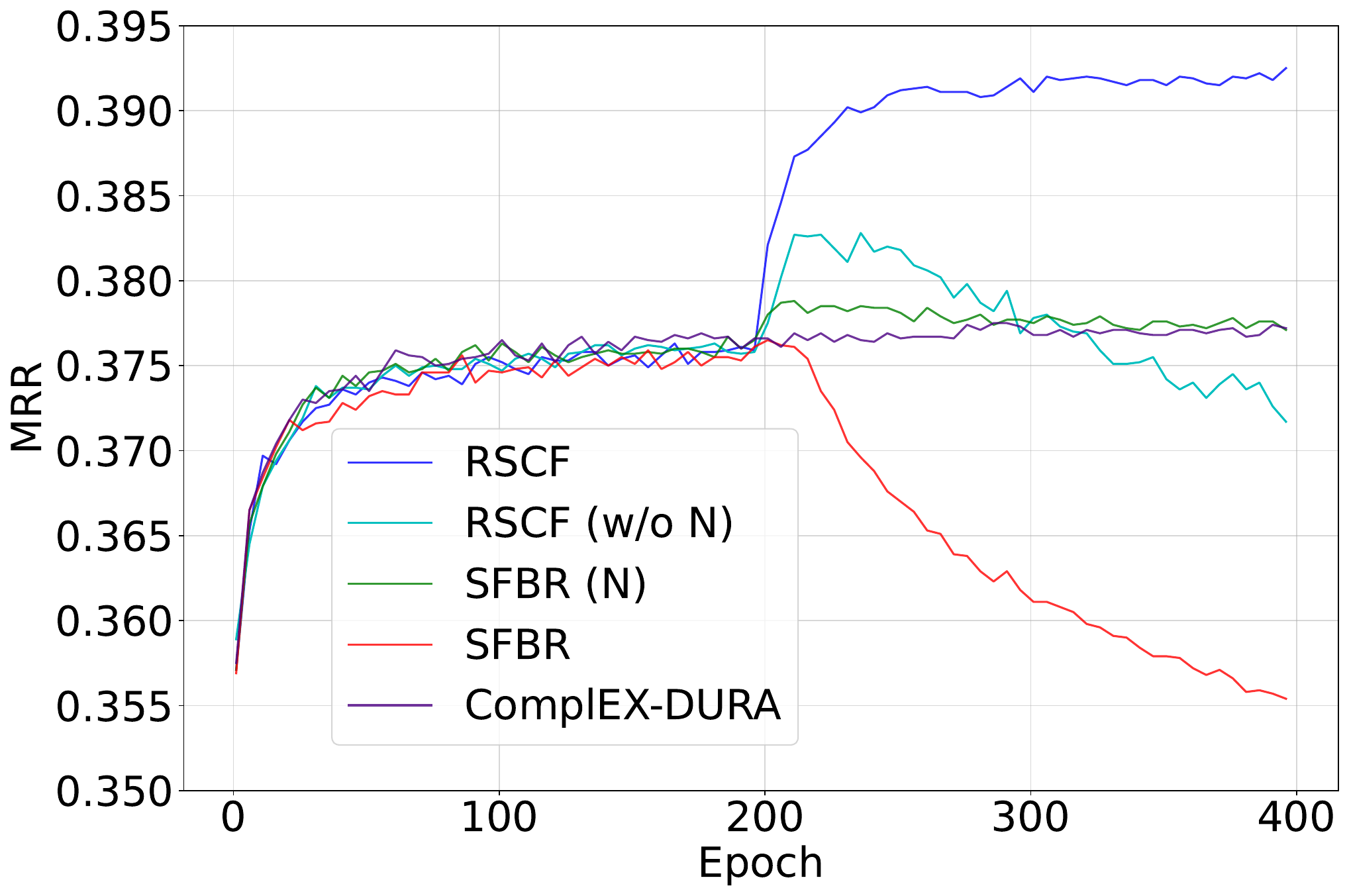}
\caption{MRR changes over epochs of RSCF, RSCF (w/o N), SFBR (N), SFBR, and ComplEX on FB15k-237.}
\label{fig:MRR}
\end{figure}

\paragraph{Ablation Study}
Table~\ref{tab:ablation} presents ablation studies of RSCF to verify the effectiveness of each component. \textcircled{a} and \textcircled{b} \& \textcircled{c} are the components of ET described in Equation~\ref{eq:main_method_final}. In the ablation study, \textcircled{b} \& \textcircled{c} are combined because \textcircled{b} \& \textcircled{c} should be used simultaneously to maintain the original scale. In Table~\ref{tab:ablation}, RSCF shows higher performance compared to the other ablated models in both TransE and ComplEX, suggesting that each component of RSCF contributes to the effectiveness of RSCF.  Especially, Figure~\ref{fig:MRR} shows that w/o normalization (\textcircled{b} \& \textcircled{c}) can significantly reduce model performance and w/ normalization maintain model performance in both RSCF and SFBR in ComplEX, indicating that normalization is necessary to maintain the performance of models that use DURA regularizer.\footnote{Detailed description of SFBR (N) is presented in Appendix~\ref{appendix: SFBR (N)}.} In addition, please note that RT can reduce the performance of ComplEX-RSCF on FB15k-237 because of context information restriction.%~\ref{appendix: SFBR (N)}}.

\begin{figure}[t]
\centering

\includegraphics[width=0.9\columnwidth]{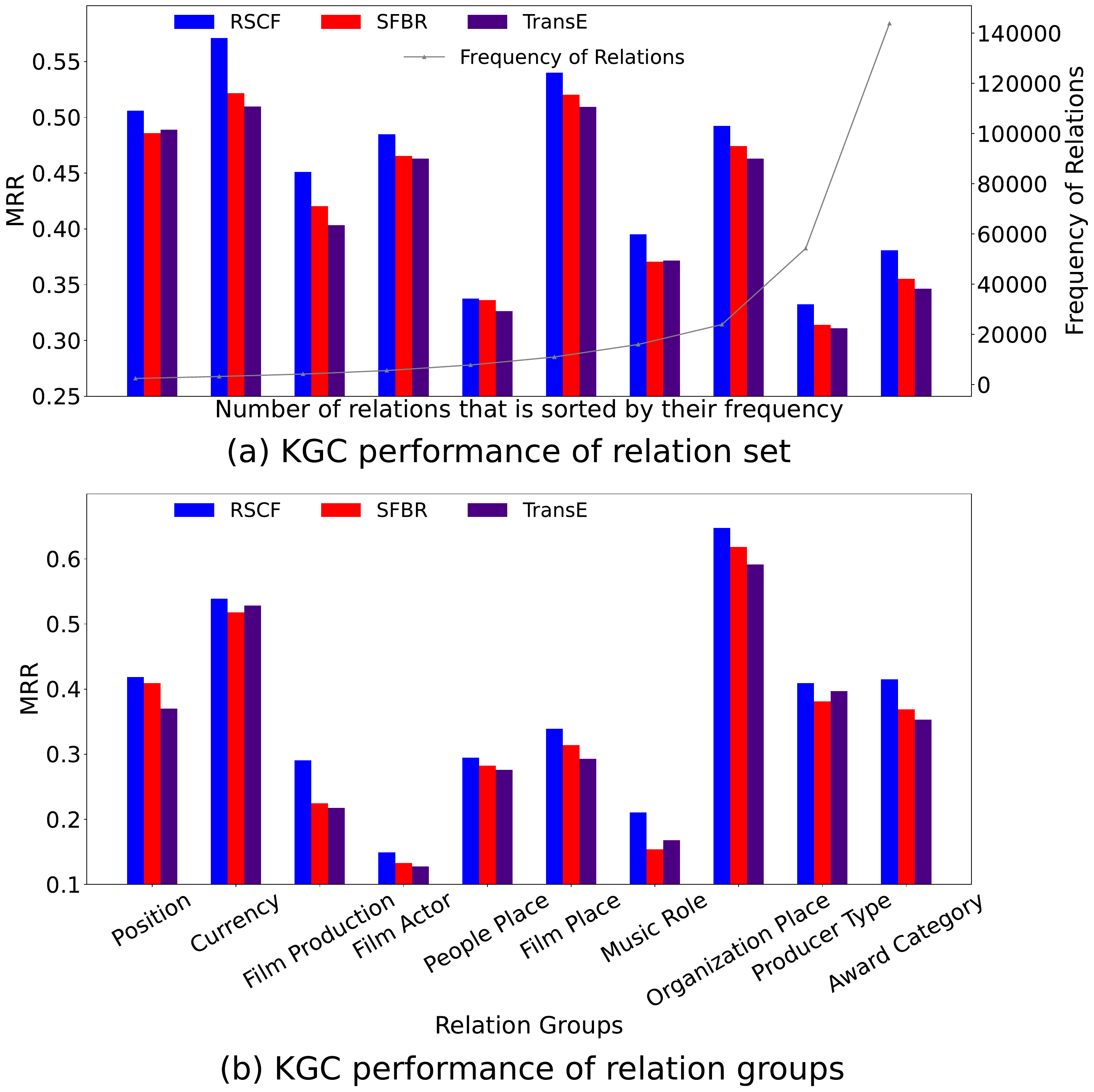}
\caption{KGC performance of the relation set that is sorted by their frequency (above) and groups of semantically similar relations observed in Figure~\ref{fig:transformation-head} (e) (below) on FB15k-237
}
\label{fig:frequency & 10_group}
\end{figure}

\paragraph{Performance on Relation Frequency and Semantically Distinguished Relation Groups}
To demonstrate the generality of applying RSCF regardless of relation frequency, we sort relations by their frequency and divide them into ten sets. Each set has an equal number of relations. Figure~\ref{fig:frequency & 10_group} above shows the MRR for each set in TransE, RSCF, and SFBR. The results shows that RSCF outperform SFBR and TransE in all sets, demonstrating the robustness of RSCF to relation frequency and showing that RSCF can be applied without trade-off between high and low frequency of relations.

Figure~\ref{fig:frequency & 10_group} below shows the MRR for each relation group as defined in Figure~\ref{fig:transformation-head} (e). RSCF outperformed SFBR and TransE in all groups, demonstrating that RSCF can be utilized without specific bias to the semantics of relations and that incorporating relation semantics into the transformation function can improve model performance.

\paragraph{Qualitative Example Analysis} For qualitative analysis, Table~\ref{tab:positive example} presents sampled queries, their correct answers, related triples with the sample queries, and the ranks obtained by RSCF and SFBR. Relations in sample queries and related triples belong to the same relation group (people place). In Table~\ref{tab:positive example}, RSCF shows enhanced performance compared to SFBR, indicating that RSCF can use trained bias between semantically similar relations.

\subsection{In-Depth Analysis}
\paragraph{Relation-Semantics Consistency of ET and EE}
Figure~\ref{fig:transformation-head} shows ET and their corresponding EE of SFBR and RSCF via T-SNE. RSCF represents a more concentrated cluster compared to SFBR, which indicates that similar relations have similar ET and EE in RSCF; in other words, RSCF satisfies relation-semantic consistency.

\begin{figure}
\centering

\includegraphics[width=0.48\textwidth]{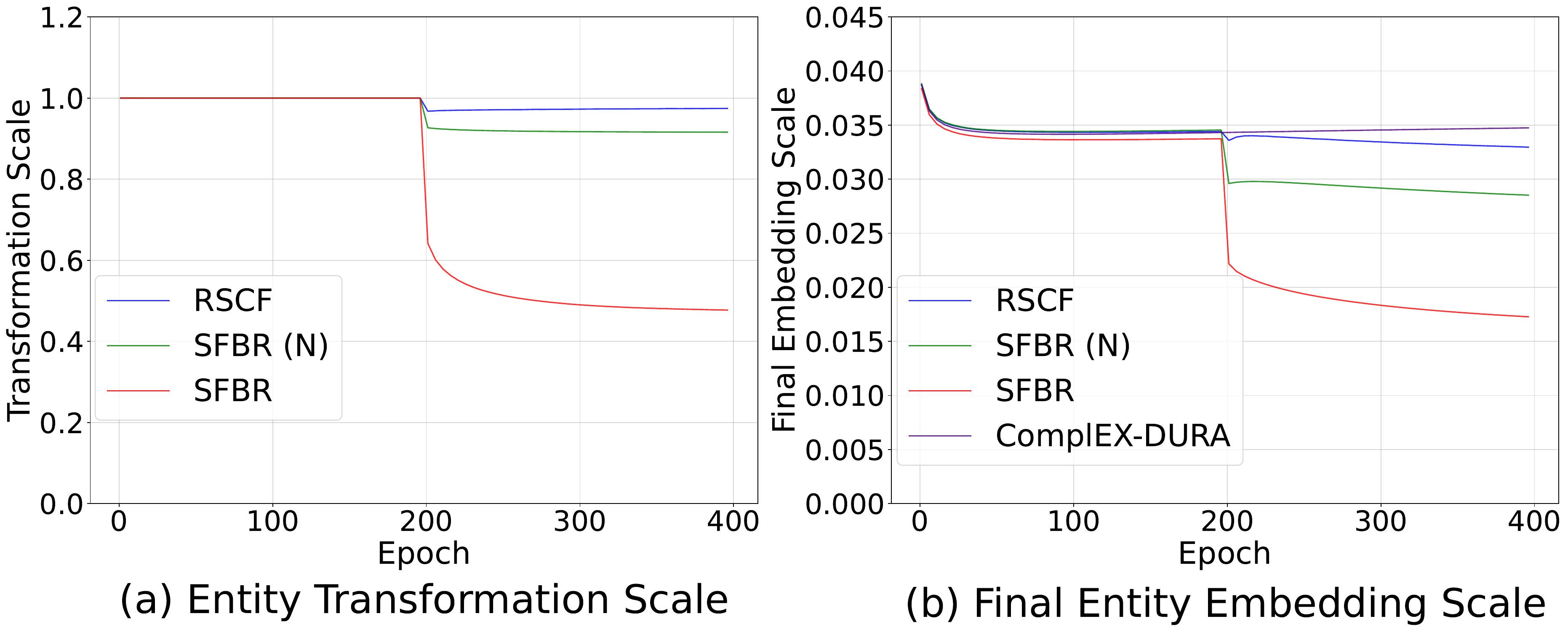}
\caption{Entity transformation scale (left) and final entity embedding scale (right) of RSCF, SFBR (N), SFBR, and ComplEX-DURA over epochs on FB15k-237. DURA is applied in all epochs and RSCF and SFBR are applied after 200 epochs. 
}
\centering
\label{fig:TS, FES}
\end{figure}
\paragraph{Recovery of Embedding Scale and Score Distribution}

Figure~\ref{fig:TS, FES} presents transformation scale and final entity embedding scale over epochs on FB15k-237, using ComplEX as the baseline. Following the approach of SFBR, DURA is applied in all epochs, and RSCF and SFBR are plugged in after 200 epochs. The results show that SFBR decreases both transformation scale and final entity embedding scale. In contrast, RSCF and SFBR (N) maintain scales, indicating that normalization helps preserve the embedding scale due to normalization. As shown in Figure~\ref{fig:MRR}, MRR decreases for SFBR and RSCF w/o normalization but increases for both RSCF and SFBR (N), implying that entity embedding concentration negatively impacts model performance. 

\begin{figure}[t]
    \centering    \includegraphics[width=0.9\columnwidth]{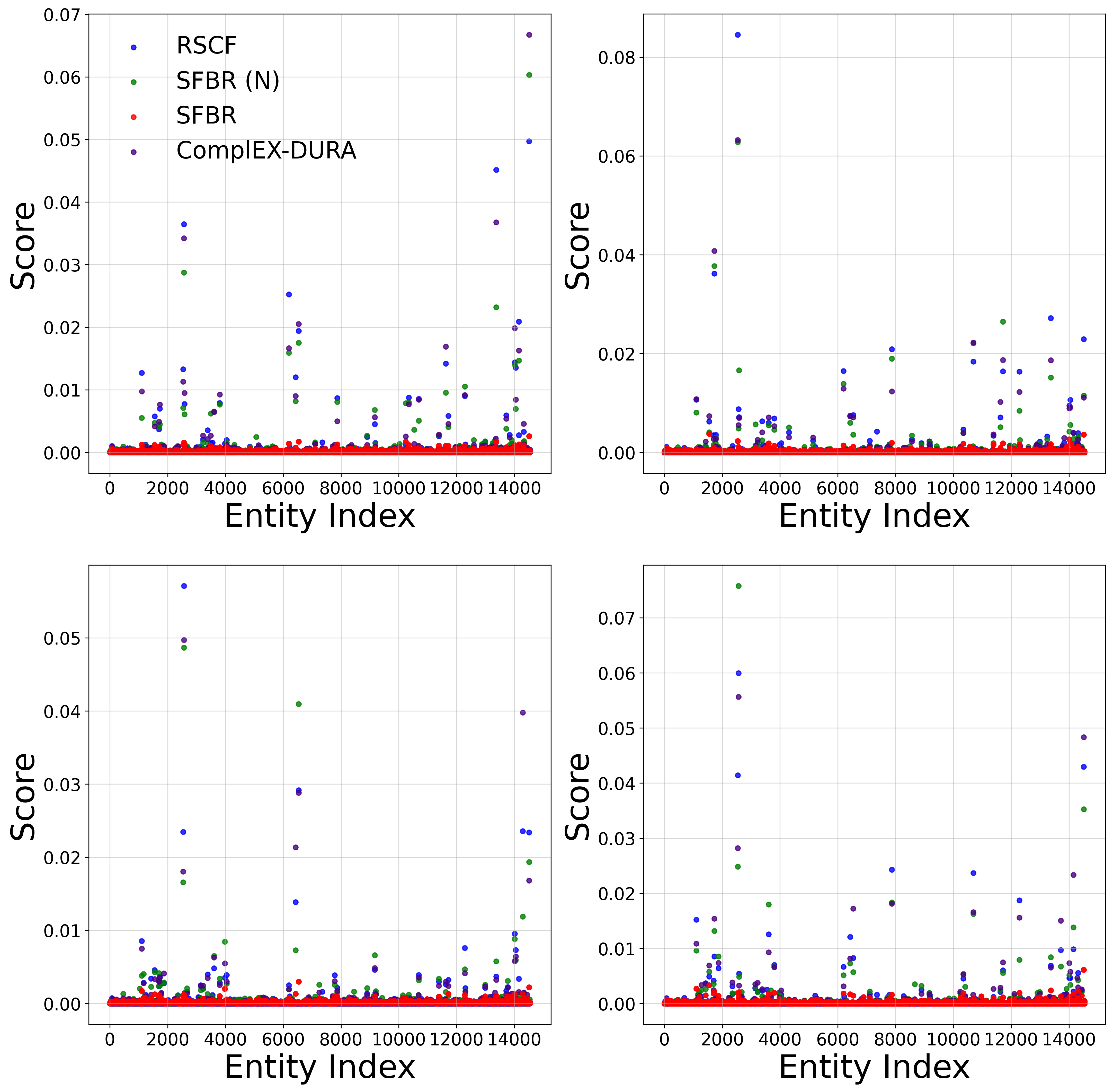}
    \caption{Score distribution of all entities for randomly selected queries from Figure~\ref{fig:Score Over Smoothing} (a)}
    \centering
    \label{fig:query graph}
\end{figure}

To investigate the detailed change of score distribution, we present the score distribution of randomly sampled queries from Figure~\ref{fig:Score Over Smoothing} (a) in Figure~\ref{fig:query graph}. SFBR shows near-zero scores for most entities, with significantly similar distributions across queries. However, by applying normalization or using RSCF, the diversity of scores is recovered as the original base model.

\paragraph{Performance Decrease by Entity Embedding Concentration}

To assess the impact of indistinguishable score distribution, we conduct a performance evaluation for the selected relation that shows critical entity embedding concentration in Figure~\ref{fig:Score Over Smoothing}. 
Table~\ref{tab:Score Over Smoothing MRR} presents the MRR for all queries associated with the selected relation. 
SFBR shows significantly lower performance than RSCF, SFBR (N), and the ComplEX. This result implies that indistinguishable score distribution strongly affects the prediction of SFBR, and simply applying normalization can recover it. 

\begin{table}[t]
\centering
\footnotesize
\scriptsize
\begin{tabular}{lllc}
\hline
Model                 & MRR                      & H@10       & Concentration     \\ \hline
ComplEX-RSCF     &\textbf{.375}  & \textbf{.609} & \xmark \\
ComplEX-SFBR (N)   &.366                   & .587          & \xmark   \\
ComplEX-SFBR        &.267                  & .522     & \cmark \\
ComplEX-DURA            & .347                  & .609   & \xmark       \\ \hline
\end{tabular}
\caption{KGC performance of all queries associated with the relation that shows strong concentration of entity embedding in SFBR. Concentration presents entity embedding concentration.}

\label{tab:Score Over Smoothing MRR}
\end{table}
\section{Conclusion}
In this paper, we address the limit in inducing relation-semantics consistency, implying that semantically similar relations have similar entity transformation, on entity transformation models for KGC, especially SFBR. 
We clarify two causes, disconnected entity transformation representation and entity embedding concentration, and provide a novel relation-semantics consistent filter (RSCF) method. Its entity transformation use shared affine transform to generate the change of entity embedding, normalize it and add it to the embedding for enhancing the semantic consistency. Also, RSCF adds relation transformation and prediction for enhancing the semantics.
This method significantly improves the performance of KGC compared to state-of-the-art KGE methods for overall relations.

\section{Limitations}
RSCF uses the simplest form of affine transformation, but it has a limit of expressing all changes across all embeddings, which requires more advanced approach. Future work should extend the method to additional KGE models to enhance generality.

\section{Acknowledgements}
This work was partly supported by Institute of Information \& communications Technology Planning \& Evaluation (IITP) grant funded by the Korea government (MSIT) (No.2019-0-01842, Artificial Intelligence Graduate School Program (GIST)) and the National Research Foundation of Korea (NRF) grant funded by the Korea government (MSIT) (No.2022R1A2C2012054, Development of AI for Canonicalized Expression of Trained Hypotheses by Resolving Ambiguity in Various Relation Levels of Representation Learning)

% \section*{Acknowledgements}

% This document has been adapted
% by Steven Bethard, Ryan Cotterell and Rui Yan
% from the instructions for earlier ACL and NAACL proceedings, including those for 
% ACL 2019 by Douwe Kiela and Ivan Vuli\'{c},
% NAACL 2019 by Stephanie Lukin and Alla Roskovskaya, 
% ACL 2018 by Shay Cohen, Kevin Gimpel, and Wei Lu, 
% NAACL 2018 by Margaret Mitchell and Stephanie Lukin,
% Bib\TeX{} suggestions for (NA)ACL 2017/2018 from Jason Eisner,
% ACL 2017 by Dan Gildea and Min-Yen Kan, 
% NAACL 2017 by Margaret Mitchell, 
% ACL 2012 by Maggie Li and Michael White, 
% ACL 2010 by Jing-Shin Chang and Philipp Koehn, 
% ACL 2008 by Johanna D. Moore, Simone Teufel, James Allan, and Sadaoki Furui, 
% ACL 2005 by Hwee Tou Ng and Kemal Oflazer, 
% ACL 2002 by Eugene Charniak and Dekang Lin, 
% and earlier ACL and EACL formats written by several people, including
% John Chen, Henry S. Thompson and Donald Walker.
% Additional elements were taken from the formatting instructions of the \emph{International Joint Conference on Artificial Intelligence} and the \emph{Conference on Computer Vision and Pattern Recognition}.

\bibliography{custom}

\appendix

\section{Appendix A}
\label{appendix: A}
\subsection{Relation Groups for Entity Transformation}

\begin{figure}[h]
\centering

\includegraphics[width=0.8\columnwidth]{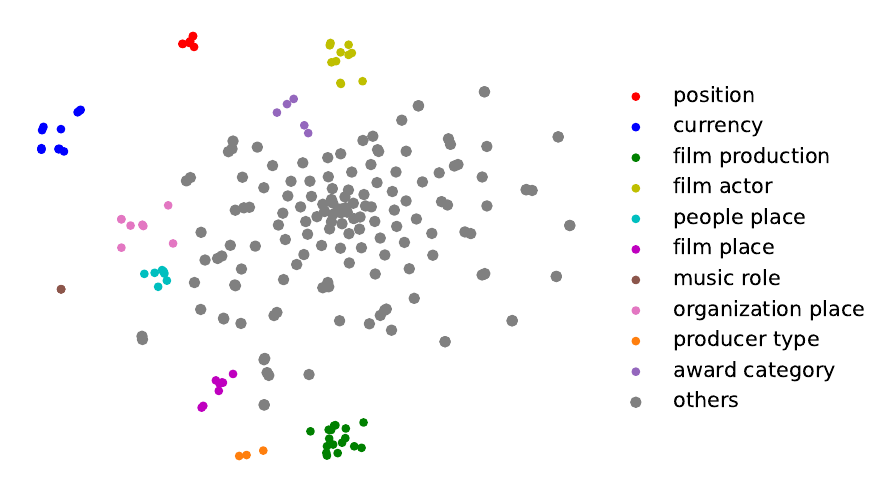}
\caption{Visualization of relation embeddings of
TransE using T-SNE} 
\centering
\label{fig:original_relation}
\end{figure}

Figure~\ref{fig:original_relation} illustrates the relation embedding of TransE. We select ten relation groups whose relation embeddings build clear and mutually decoupled clusters, which implies semantically distinguished relation groups. The other relations are plotted as grey points. The relations corresponding to each group are listed in Table~\ref{tab:relation groups}. Note that similar relations belong to the same group.

\begin{figure}[h]
\centering

\includegraphics[width=0.75\columnwidth]{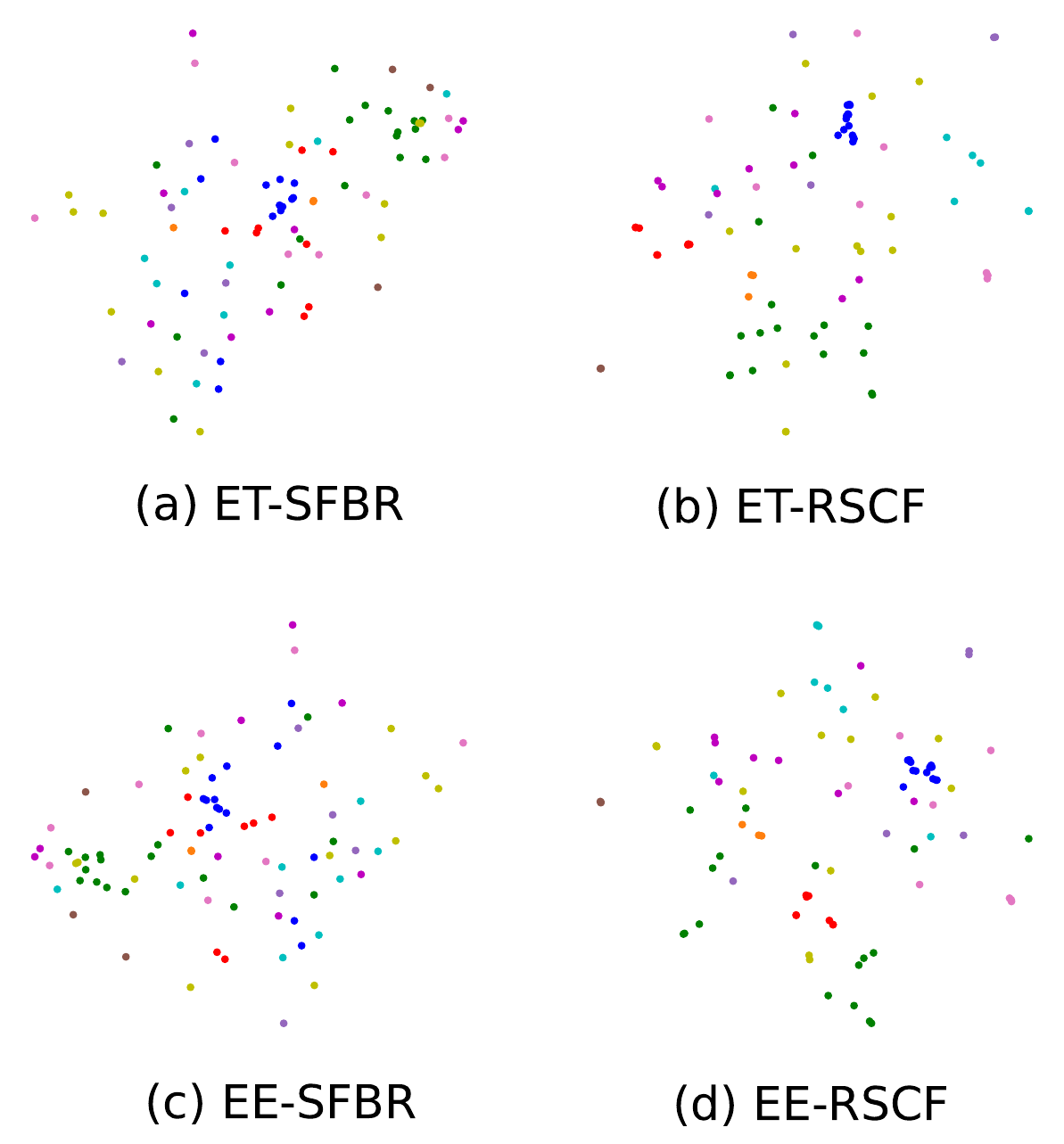}
\caption{Tail entity transformations and entity embeddings for semantically similar relation groups. (a) and (b) indicate ET of SFBR and RSCF, (c) and (d) indicate
EE of SFBR and RSCF.} 
\centering
\label{fig:transformation-tail}
\end{figure}

\begin{table}[h]
\centering
\tiny
\begin{tabular}{lcccc}
\hline
Metric                 & ET-SFBR & ET-RSCF & EE-SFBR & EE-RSCF \\ \hline
Intra Cluster Distance ($\downarrow$)    & 5.20    & 0.52    & 2.94    & 0.67    \\ 
Inter Cluster Distance ($\uparrow$) & 0.46    & 0.70     & 0.46    & 0.74    \\ \hline
\end{tabular}
\caption{Intra Cluster Distance and Inter Cluster Distance of tail entity transformation and entity embedding of SFBR and RSCF.}
\label{tab:transformation_tail_SC_ICD}
\end{table}

\subsection{Distribution of Tail Entity Transformations and Corresponding Entity Embedding} Figure~\ref{fig:transformation-tail} presents the t-SNE visualization of tail ET and corresponding EE of RSCF and SFBR. Even in the tail, RSCF shows more concentrated clusters. Also, in Table~\ref{tab:transformation_tail_SC_ICD}, RSCF exhibits lower Intra Cluster Distance and higher Inter Cluster Distance compared to SFBR. The groups for ET are equivalent to the relation groups in Figure~\ref{fig:original_relation}, and an entity for EE is randomly selected from FB15k-237. 

\subsection{Measurement of Cluster Concentration}
\label{appendix:concentraion_score}
To measure distance between elements within a cluster, we defined intra cluster distance score as follows:
\begin{align}
    \begin{split}
     \sum_k^n\sum_i^m \frac{||(k_i-C_k)||}{n||C_k||}
     \end{split}
    \label{eq:Concentration Score}
\end{align}
where $k$ is clear and mutually decoupled clusters and $k_i$ is $i$-th vector embedding of ET in group $k$ and $C_k$ is centroid of cluster $k$ that can be calculated as:
\begin{align}
    \begin{split}
     C_k = \frac{\sum_i^mk_i}{m}
     \end{split}
    \label{eq:Centroid }
\end{align}
In the equation~\ref{eq:Concentration Score}, the vector norm of $C$ ($||C||$) is used because of the relative intra cluster distance score for clusters. Also to evaluate the distance between different clusters we defined inter cluster distance score as follows:
\begin{align}
    \begin{split}
     \sum_k^n \frac{||(C_k-C_{kc})||}{\sum_i^m||k_i||}
     \end{split}
    \label{eq:Inter cluster distance}
\end{align}
where $C_{kc}$ represent the centroid that is closest to $C_k$, and it can be written as:
\begin{align}
    \begin{split}
     C_{kc} = \min_{k \neq j} \{||C_k - C_j|| \}
     \end{split}
    \label{eq:closet cluster}
\end{align}

In Equation~\ref{eq:Inter cluster distance}, the norm of the cluster, which is calculated as the sum of the elements in the cluster, is used for the relative inter cluster distance.

\begin{table*}[h]
\centering
\scriptsize
\begin{tabular}{lccccccccc}
\hline
\multicolumn{1}{c}{\multirow{2}{*}{\textbf{Knowledge Graph Embedding}}} & \multicolumn{3}{c}{\textbf{WN18RR}}                                          & \multicolumn{3}{c}{\textbf{FB15k-237}}                                       & \multicolumn{3}{c}{\textbf{YAGO3-10}}                                        \\ \cline{2-10} 
\multicolumn{1}{c}{}                                                    & \multicolumn{1}{l}{MRR} & \multicolumn{1}{l}{H@1} & \multicolumn{1}{l}{H@10} & \multicolumn{1}{l}{MRR} & \multicolumn{1}{l}{H@1} & \multicolumn{1}{l}{H@10} & \multicolumn{1}{l}{MRR} & \multicolumn{1}{l}{H@1} & \multicolumn{1}{l}{H@10} \\ \hline
TuckER~\cite{balavzevic2019tucker}                                                                  & .470                    & .443                    & .526                     & .358                    & .266                    & .544                     & -                       & -                       & -                        \\
QuatE~\cite{zhang2019quaternion}                                                                   & .488                    & .438                    & .582                     & .348                    & .248                    & .550                     & -                       & -                       & -                        \\ 
MuRP~\cite{balazevic2019multi}                                                                    & .481                    & .440                    & .566                     & .335                    & .243                    & .518                     & -                       & -                       & -                        \\
HAKE~\cite{zhang2020learning}                                                                    & \underline{.497}                    & .452                    & .582                     & .346                    & .250                    & .542                     & .545                    & .462                    & .694                     \\
RoTH~\cite{chami2020low}                                                                    & .496                    & .449                    & \underline{.586}                     & .344                    & .246                    & .535                     & .570                    & .495                    & .706                     \\
DualE~\cite{cao2021dual}                                                                   & .492                    & .444                    & .584                     & .365                    & .268                    & .559                     & -                       & -                       & -                        \\
FieldE~\cite{nayyeri2021knowledge}                                                                  & .48                     & .44                     & .57                      & .36                     & .27                     & .55                      & .51                     & .41                     & .68                      \\
Rot-Pro~\cite{song2021rot}                                                                 & .457                    & .397                    & .577                     & .344                    & .246                    & .540                     & .542                    & .443                    & .699                     \\

HAKE-CAKE~\cite{niu2022cake}                                                               & -                       & -                       & -                        & .321                    & .226                    & .515                     & -           & -              & -             \\

GIE~\cite{yang2022knowledge}                                                               & .491                       & .452                       & .575                        & .362                    & .271                    & .552                     & .579              & .505              & .709               \\
ComplEX-ER~\cite{cao2022er}                                                               & .494                       & \underline{.453}                       & .575                        & \underline{.374}                    & \underline{.282}                    & \underline{.563}                     & \underline{.588}              & \underline{.515}              & \textbf{.718}               \\

TranSHER~\cite{li2022transher}    & -                      & -                       & -                        & .360                    & .264                    & .551                     & -              & -              & -               \\

STaR-DURA~\cite{li2022star}    & \underline{.497}                      & .452                       & .583                        & .368                    & .273                    & .557                     & .585              & .513              & \underline{.713}               \\
\hline
{\multirow{2}{*}{ComplEX-DURA-RSCF (Ours)}}                                                            & \textbf{.503}       & \textbf{.460}       & \textbf{.588}      & \textbf{.388} & \textbf{.295} & \textbf{.573} & \textbf{.589} & \textbf{.516} & \textbf{.718}  \\ 
    & $\pm$.001       & $\pm$.001       & $\pm$.002      & $\pm$.001 & $\pm$.002 & $\pm$.004 & $\pm$.002 & $\pm$.003 & $\pm$.002  \\
\hline
\end{tabular}
\caption{Test performance of KGE-based KGC on FB15k-237, WN18RR and YAGO3-10. Bold indicates the best result, and underlined signifies the second best result.}
\label{tab:KGE-comparison}
\end{table*}

\begin{table*}[h]
    \centering
    \scriptsize
    \begin{tabular}{cccccccc}
    \hline
    \multirow{2}{*}{\textbf{Dataset}} &
    \multirow{2}{*}{\textbf{Entities}} &
    \multirow{2}{*}{\textbf{Relations}} &
    \multirow{2}{*}{\textbf{Entities/Relations}} &
    \multirow{2}{*}
    {\textbf{Triples/Relations}} &
    
    \multicolumn{3}{c}{\textbf{Triples}}\\
        \cline{6-8}
    & & & & & \textbf{Train} & \textbf{Valid} & \textbf{Test} \\
    \hline
    WN18RR & 40,943 & 11 & 3,722 & 7,894 & 86,835 & 3,034 & 3,134 \\
    FB15k-237 & 14,541 & 237 & 61  & 1,148 & 272,115 & 17,535 & 20,466 \\
    YAGO3-10 & 123,182 & 37 & 3,329 & 29,163 & 1,079,040 & 5,000 & 5,000\\
    \hline
    \end{tabular}
    % }
    \caption{Statistics of KGC Benchmark Datasets}
    \label{tab:Dataset}
\end{table*}

\section{Appendix B}
\label{appendix: B}
\subsection{Proof of scale decrease of ET}
\label{appendix: Proof of scale decrease of ET}
Let $\mathsf{W_{r_j}}$ is the ET of SFBR and $w_{r_j, n}$ is $n$-th element of $\mathsf{W_{r_j}}$, than the gradient of $w_{r_j, n}$ in DURA can be calculated as:
\begin{align}
    \begin{split}
    &\sum_{p}\frac{dL}{dw_{r_j,n}}||\mathsf{w_{r_j,n}h_{i, n} {r_{j,n}}}||^2_2  + ||\mathsf{w}_{r_jn}\mathsf{h_{i,n}}||^2_2 \\ =
     &\sum_{p}\frac{dL}{dw_{r_j,n}}\mathsf{w_{r_j,n}^2(h_{i, n} {r_{j,n}}})^2  + \mathsf{w}_{r_jn}^2\mathsf{h_{i,n}}^2 \\ =
    &\sum_{p}2\mathsf{w_{r_j,n}(h_{i, n} {r_{j,n}}})^2  + 2\mathsf{w}_{r_jn}\mathsf{h_{i,n}}^2
    \end{split}
    \label{eq:scale decrease of ET}
\end{align}
The gradient of ET shows that the gradient of $w_{r_j, n}$ has always same sign with $w_{r_j, n}$ parameters. Therefore, gradient descent always reduces the scale of the parameters regardless of their sign.

\subsection{Normalization of Change for Reducing Entity Embedding Concentration}
\label{appendix: Normalization of Change for Reducing Entity Embedding Concentration}
The change generated from the affine transformation is normalized by its length, expressed as $\mathbf{N}_p\mathbf{(rA)}$ in the part \textcircled{b}. This normalization alleviates critical entity embedding concentration via reducing scale decrease of transformed entity embeddings $\mathbf{e_r}$ in DURA regularization.
In our relation specific rooted ET, the change of $\mathbf{e_r}$ is simply written as 

\begin{align}
    \begin{split}
    \lVert \mathbf{\alpha} \otimes \mathbf{e} \rVert_{p}
    \end{split}
    \label{eq:transformation_changin rate}
\end{align}
where $\mathbf{\alpha} = \mathbf{N}_p\mathbf{(rA)}$. 
This value has a maximum when $\alpha$ has the same direction to $\mathbf{e}$. Since $\mathbf{\alpha}$ is a unit vector in p-norm, $\mathbf{\alpha} = \mathbf{e}/\lVert \mathbf{e} \lVert_{p}$ . Then, the maximum change is
\begin{align}
    \begin{split}
    \lVert \frac{\mathbf{e}}{\lVert \mathbf{e} \rVert_{p}} \otimes\mathbf{e} \rVert_{p} =
    \lVert \frac{\mathbf{e}^2}{\lVert \mathbf{e} \rVert_{p}} \rVert_{p} =
    \frac{\lVert \mathbf{e}^2 \rVert_{p}}{\lVert \mathbf{e} \rVert_{p}}
    \end{split}
    \label{eq:transformation_changing rate}
\end{align}
In practice, the elements of embedding vectors are much less than 1 in most cases. Therefore, the maximum change $\lVert \mathbf{e}^2 \rVert_{p}/\lVert \mathbf{e} \rVert_{p}$ is significantly lower than the unrestricted scale change in SFBR.  

\subsection{
Empirical Experiments on Maintaining Consistency through Monte Carlo Simulation}
\label{appendix: Monte Carlo}
Linear transformation ensures consistency when relation embeddings exist on a line. Furthermore, for relations that do not lie on the line, we can predict that that similar relations will have similar ETs because of the continuous property of linear transformation, i.e. if $r_1 \approx r_2$ then $r_1A \approx r_2A$.
However, there has been no research on the proportion of these relations that maintain consistency after transformation and also, after normalization, and it is extremely challenging to determine this proportion through mathematical formulations. Therefore, we conducted an empirical analysis using Monte Carlo simulations to investigate the consistency of points that do not lie on the line. Table~\ref{tab:Monte-carlo simulation} presents the proportion of consistency maintained under various conditions based on Monte Carlo simulations. For the experiment, we divided the scenarios into four cases: (1) when three randomly generated points (A, B, C) lie on the same line  and the distance between A and C is greater than the distance between A and B (Line), (2) when the three points (A, B, C) do not lie on the line and the distance between A and C is greater than the distance between A and B$(\frac{|\overline{AC}|}{|\overline{AB}|} > 1)$, (3) when the distance between A and C is at least 1.01 times greater than the distance between A and B $(\frac{|\overline{AC}|}{|\overline{AB}|} > 1.01)$, and (4)  when the distance between A and C is at least 1.02 times greater than the distance between A and B $(\frac{|\overline{AC}|}{|\overline{AB}|} > 1.02)$ and sampling was performed 10,000 times for each condition. Under each condition, we measured the proportion of cases where $(|\overline{AC}| > |\overline{AB}| )$ was maintained even after Transformation (\textcircled{a}), Normalization (\textcircled{b}), and the Add one (\textcircled{c}). The results showed that consistency was preserved in most cases, even when the three points did not lie on the same line. Specifically, in condition where $(\frac{|\overline{AC}|}{|\overline{AB}|} > 1)$, over 72.8\% of the samples maintained consistency in Transformation (\textcircled{a}). Furthermore, in cases where $(\frac{|\overline{AC}|}{|\overline{AB}|} > 1.02)$ approximately 87.5\% of the samples maintained consistency after Transformation (\textcircled{a}). Also in Normalization (\textcircled{a}), over 86.9\% of the samples maintained consistency in $(\frac{|\overline{AC}|}{|\overline{AB}|} > 1)$ and 99.4\% in $(\frac{|\overline{AC}|}{|\overline{AB}|} > 1.02)$. Considering that relations tend to form clusters based on their semantics because of score function and RP~\cite{chen2021relation}, it is difficult to say that these conditions are unrealistic, indicating that our method performs robustly across various conditions.

\section{Appendix C}

\label{appendix: C}
\subsection{Datasets}
\label{appendix: dataset statistics}
We evaluate the RSCF using three widely-used
datasets: WN18RR, FB15k-237, and YAGO3-10. WN18RR, FB15k-237 and YAGO3-10 are subsets of WN18~\cite{bordes2013translating}, FB15k~\cite{bordes2013translating}, and YAGO3~\cite{mahdisoltani2013yago3}, respectively, designed to alleviate the test set leakage problem. Statistics of these
datasets are shown in Table~\ref{tab:Dataset}.

\begin{table*}[t]
\centering
\small
\begin{tabular}{lcccccc}
\hline
\multicolumn{1}{c}{\multirow{2}{*}{\textbf{\begin{tabular}[c]{@{}c@{}}Distance-Based Model\\ with Entity Transformation\end{tabular}}}} & \multicolumn{3}{c}{\textbf{WN18RR}}                                   & \multicolumn{3}{c}{\textbf{FB15k-237}}        \\ \cline{2-7} 
\multicolumn{1}{c}{}                                                                                                                    & MRR                   & H@1                   & H@10                  & MRR           & H@1           & H@10          \\ \hline
TransE-SFBR (Diag)~\cite{liang2021semantic}     & .242 & .028 & .548   & .338          & .240 & .538          \\
TransE-SFBR (Linear-2)~\cite{liang2021semantic}          & .263   & .110    & .495  & .354  & .258  & .545   \\
RotatE-SFBR (Diag)~\cite{liang2021semantic}            & .489         & .437                  & \textbf{.593}         & .351          & .254          & .549 \\

RotatE-SFBR (Linear-2)~\cite{liang2021semantic}             & .490                  & .447                  & .576                  & .355          & .258          & .553 \\
\hline

{\multirow{2}{*}{TransE-RSCF}}      & .267         & .066         & .546         & .363 & \underline{.264} & \textbf{.558} \\ 

& 	$\pm$.001       & 	$\pm$.002       & 	$\pm$.002      & 	$\pm$.001 & 	$\pm$.001 & 	$\pm$.001 \\

{\multirow{2}{*}{TransE-RSCF (Linear-2)}}   & .343         & .232         & .499         & .359 & .262 & .552 \\ 
& 	$\pm$.009       & 	$\pm$.014       & 	$\pm$.003      & 	$\pm$.001 & 	$\pm$.001 & 	$\pm$.001 \\

{\multirow{2}{*}{RotatE-RSCF}}   & \underline{.493}         & \underline{.447}         & \underline{.584}                  & \underline{.363} & \textbf{.268} & \underline{.556}          \\ 
& 	$\pm$.001       & 	$\pm$.001       & 	$\pm$.001      & 	$\pm$.000 & 	$\pm$.001 & 	$\pm$.001 \\

{\multirow{2}{*}{RotatE-RSCF (Linear-2)}}    & \textbf{.495}         & \textbf{.452}         & .578         & \textbf{.364} & \textbf{.268} & \underline{.556}          \\ 
& 	$\pm$.001       & 	$\pm$.001       & 	$\pm$.001      & 	$\pm$.000 & 	$\pm$.000 & 	$\pm$.001 \\
\hline
\end{tabular}%
 
\caption{Test performance of DBM-based RSCF and SFBR on FB15k-237 and WN18RR. Bold indicates the best result, and underlined signifies the second best result.}
\label{tab:DBM with ET}
\end{table*}

\begin{table*}[t]
\centering
\scriptsize
\begin{tabular}{lccccccccc}
\hline
\multicolumn{1}{c}{\multirow{2}{*}{\textbf{\begin{tabular}[c]{@{}c@{}}Tensor Decomposition Model\\ with Eentity Transformation\end{tabular}}}} & \multicolumn{3}{c}{\textbf{WN18RR}}           & \multicolumn{3}{c}{\textbf{FB15k-237}}        & \multicolumn{3}{c}{\textbf{YAGO3-10}}         \\ \cline{2-10} 
\multicolumn{1}{c}{}                                                                                                                           & MRR           & H@1           & H@10          & MRR           & H@1           & H@10          & MRR           & H@1           & H@10          \\ \hline

CP-DURA + SFBR~\cite{liang2021semantic}                                                                                                                             & .485          & .447          & .561          & .370 & .274 & .563          & .582          & .510          & .711          \\

RESCAL-DURA + SFBR~\cite{liang2021semantic}                                                                                                                        & .500 & .458 & .581 & .369 & .276 & .555 & .581 & .509 & .712 \\

ComplEX-DURA + SFBR~\cite{liang2021semantic}                                                                                                                        & .498          & .454          & \underline{.584}          & .374 & .277 & \underline{.567}          & .584          & .512          & .712          \\ \hline

{\multirow{2}{*}{CP-DURA + RSCF}}                                                                                                                                  & .486 & .447 & .561 & .379 & .287          & .565 & \underline{.585}& \underline{.514} & .711 \\ 
& 	$\pm$.001       & 	$\pm$.001       & 	$\pm$.001      & 	$\pm$.000 & 	$\pm$.000 & 	$\pm$.001& 	$\pm$.000 & 	$\pm$.001 & 	$\pm$.001 \\

{\multirow{2}{*}{RESCAL-DURA + RSCF}}                                                                                                                              & \textbf{.507} & \textbf{.467} & .581 & \underline{.381} & \underline{.289} & .562 & .584 & .511          & \underline{.716} \\ 
& 	$\pm$.000       & 	$\pm$.000       & 	$\pm$.000      & 	$\pm$.000 & 	$\pm$.000 & 	$\pm$.000& 	$\pm$.000 & 	$\pm$.000 & 	$\pm$.000 \\

{\multirow{2}{*}{ComplEX-DURA + RSCF}}                                                                                                                            & \underline{.503} & \underline{.460} &  \textbf{.588} & \textbf{.388} & \textbf{.295} & \textbf{.573} &  \textbf{.589} & \textbf{.516} & \textbf{.718} \\ 
    & $\pm$.001       & $\pm$.001       & $\pm$.002      & $\pm$.001 & $\pm$.002 & $\pm$.004 & $\pm$.002 & $\pm$.003 & $\pm$.002   \\
\hline
\end{tabular}%
 
\caption{Test performance of TDM-based RSCF and SFBR on FB15k-237, WN18RR, and YAGO3-10. Bold indicates the best result, and underlined signifies the second best result.}
\label{tab:TDM with ET}
\end{table*} 

\subsection{Additional Performance Experiments}
\label{appendix: Performance Compariosn of RSCF}
\begin{figure}[t]
\centering
\includegraphics[width=0.85\columnwidth]{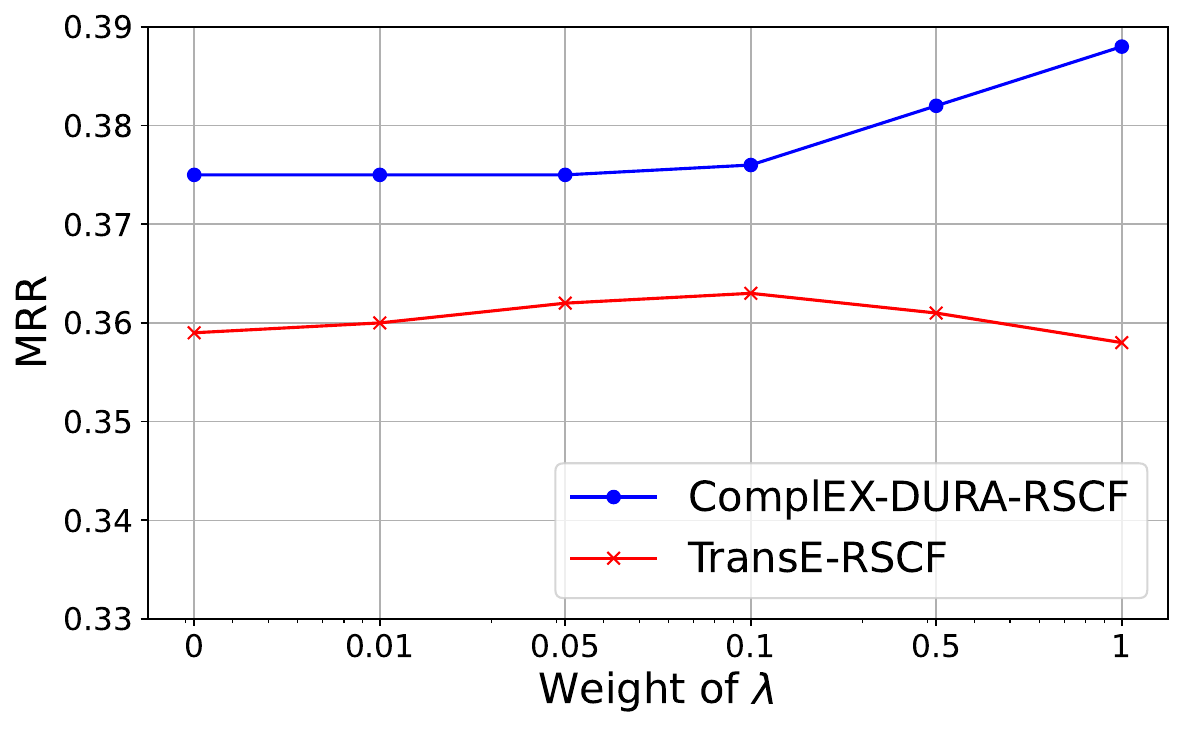}
\caption{MRR variations with respect to the RP weight $\lambda$, which is defined in Equation~(\ref{eq:RP_equation}). The performance is measured on FB15k-237.}
\centering
\label{fig:lambda_weight}
\end{figure}

\paragraph{Performance Comparison of RSCF and Other KGE Models} Table~\ref{tab:KGE-comparison} shows the performance comparison of the RSCF and previous KGE-based models on WN18RR, FB15k-237 and YAGO3-10. Overall, ComplEX-DURA+RSCF shows higher performance than other KGE models in all settings, demonstrating that the effectiveness of the
RSCF for the KGC task.
\paragraph{Performance Comparison of RSCF and SFBR} Table~\ref{tab:DBM with ET} shows the performance comparison of the DBM-RSCF and DBM-SFBR on WN18RR and FB15k-237. Overall, DBM-RSCF shows similar or higher performance than DBM-SFBR in most settings. Table~\ref{tab:TDM with ET} shows the performance comparison in TDMs. Compared to TDM-SFBR, TDM-RSCF shows consistent performance improvements in all datasets and settings.
\paragraph{Performance Comparison of RSCF with Different RP Weights} 
Figure~\ref{fig:lambda_weight} shows the MRR variations with respect to the RP weight, $\lambda$. TransE-RSCF achieves better performance at $\lambda=0.1$, while ComplEX-DURA-RSCF performs better at $\lambda=1$. In general, the MRR tends to decrease as $\lambda$ becomes smaller, suggesting that the RP plays a crucial role in enhancing model performance.

\begin{table*}[h]
\centering
\tiny
\begin{tabular}{lcccccccc}
\hline
\multicolumn{1}{c}{\multirow{2}{*}{Model}} & \multicolumn{2}{c}{Training Time} & \multicolumn{2}{c}{Inference Time} & \multicolumn{2}{c}{\# Params}  & \multicolumn{2}{c}{MRR}\\ \cline{2-9}
\multicolumn{1}{c}{}                       & WN18RR         & FB15k-237      & WN18RR         & FB15k-237  & WN18RR         & FB15k-237       &      WN18RR & FB15k-237                      \\ \hline
TransE                                     & 45m             & 1h       & 30s   &  1m 20s      & 20.48M & 14.78M        &.226 & .294                \\
PairRE                                     & -             & 3h           & -  & 1m 20s      &-  &22.52M            & - & .351             \\
T-SFBR                                & 1h             & 1h 15m  & 30s   &  1m 30s   &   20.49M & 15.25M          & .242 & .338               \\
CompoundE                                  & 2h 40m             & 2h 20m              & 30s  & 1m 20s   & 20.5M     & 9.58M  & .491 & .357                   \\ \hline
T-RSCF                                       & 3h 10m             & 5h 30m            & 30s  & 1m 50s     & 21.48M       & 18.78M & .267 & .363             \\
T-RSCF$_{\text{small}}$                                       &  1h 40m            & 3h            & 30s  & 1m 20s     & 10.49M       & 8.4M    & .263 & .358               
      \\\hline
\end{tabular}
\caption{Training time, inference time, number of parameters and MRR of RSCF and ETMs, T-SFBR (Diag) and T-RSCF indicate TransE-SFBR and TransE-RSCF, respectively. T-RSCF$_{\text{small}}$ has half of the entity/relation embedding dimension compared to T-RSCF. Inference Time denotes inference time on test set. Both training and inference time are measured on RTX3090.}
\label{tab:time complexity}
\end{table*}

\subsection{SFBR with Normalization}
\label{appendix: SFBR (N)}
To prevent entity embedding concentration, We apply normalization to SFBR that is presented as SFBR (N). Let $\mathbf{W_r}$ is relation specific ET using separate parameters, then SFBR with normalization can be written as:  
\begin{align}
    \begin{split}
    \mathbf{N}_p(\mathbf{W_r}) + 1
    \end{split}
    \label{eq:SFBR with N}
\end{align}
where $\mathrm{N}_p(\mathbf{W_r}) = \frac{\mathbf{W_r}}{\lVert \mathbf{W_r}\rVert_{p}}$.
Additionally, transformed entity embedding can be described as:
\begin{align}
    \begin{split}
    \mathbf{e_r} = (\mathbf{N}_p(\mathbf{W_r}) + 1)\mathbf{e}
    \end{split}
    \label{eq:SFBR_final}
\end{align}
where $\mathbf{e}$ is a original entity embedding.

\subsection{Extension of RSCF}
The shared affine transformation can be easily extended to $Linear-2$ that is introduced in SFBR by extending shared affine transformation $\mathbf{W_e} \in \mathbf{R}^{n \times n}$ to $\mathbf{W_e} \in \mathbf{R}^{n \times 2n}$. Therefore, RSCF (Linear-2) can be written as: 
\begin{eqnarray}
    \mathbf{W_r}^{\text{Linear-2}} &=&  
        \begin{bmatrix}
        diag(\mathbf{w}_1) & diag(\mathbf{w}_2) \\
        diag(\mathbf{w}_3) & diag(\mathbf{w}_4) \\
        \end{bmatrix}
\label{eq:sfbr_adaptation}
\end{eqnarray} 
where $\mathbf{W_r}^{\text{Linear-2}}$ $\in \mathbf{R}^{n\times n}$ is ET built from the relation specific change vector $\mathbf{N}_p(\mathbf{rA}) + 1$ of RSCF that is notated as concatenation of diagonal values of $\mathbf{w}_1, \mathbf{w}_2, \mathbf{w}_3, \mathbf{w}_4 \in \mathbf{R}^{n / 2}$.

\subsection{Computational Complexity}
\label{appendix: Complexity Analsis}
Table~\ref{tab:time complexity} presents the complexity comparison between RSCF and other ETMs. In general, because the number of parameters in KGE methods is significantly influenced by the number of entities and relations, the parameter difference between RSCF and ETMs is marginal. Additionally, although RSCF requires more training time compared to other models, considering that the proposed KGE models assume an offline learning setting and have similar inference times to RSCF, this is not a significant drawback.
Moreover, T-RSCF$_{\text{small}}$ that has half of the entity/relation embedding dimension compared to T-RSCF, exhibits higher performance than base models like TransE and T-SFBR. Especially in FB15k-237, T-RSCF$_{\text{small}}$ outperforms all other models except for T-RSCF, while requiring lower computational cost than T-RSCF. These results show that RSCF can be effective even in resource-constrained situations.

\subsection{Implementation Details}
\label{appendix:Implementation Details}
When training the RSCF, we followed the experimental settings described in the SFBR~\cite{liang2021semantic}. Following the settings of SFBR, RSCF and RSCF (Linear-2) are applied to both head and tail entities in DBM, and RSCF is applied to only the head entity in TDM due to computational costs~\cite{liang2021semantic}.  For the same reason, both head and tail entities are used for RT in DBM, whereas only the head entity is used in TDM. In FB15k-237, the entity/relation ratio and the train/relation ratio are significantly lower compared to the other two datasets, which restricts the context information that each relation can obtain. This restriction is more critical in TDM, which uses only the head entity, and thus RT is not applied to TDM on FB15k-237. The hyper-parameters in DBM are consistent with the hyper-parameters in~\citet{sun2018rotate}, and hyper-parameters of TDM are consistent with the hyper-parameters in~\citet{zhang2020duality}. Additionally, similar to~\citet{chen2021relation}, we searched the weight of RP over [1, 0.5, 0.1, 0.05, 0.01, 0]. The presented results of RSCF represent the mean of the three runs for each model. Experiments for the DBM were conducted on an NVIDIA 3090 with 24GB of memory, while experiments for the TDM were conducted on an NVIDIA 2080TI with 11GB and an A100 with 40GB of memory was used for both DBM and TDM.

\section{Appendix D}
\label{appendix: D}

\subsection{Special Cases with RSCF}

\label{appendix:Special Cases with RSCF}
Let $\mathbf{h_r}$, $\mathbf{t_r}$ are transformed head and tail embedding by RSCF, then the score function $d_r(\mathbf{h}, \mathbf{r})$ of TransE-RSCF can be expressed as:
\begin{align}
    \begin{split}
     d_r(\mathbf{h}, \mathbf{r}) &= \lVert \mathbf{h_r} + \mathbf{r_{ht}} - \mathbf{t_r} \rVert
    \end{split}
    \label{eq:TransE-RSCF}
\end{align}
The score function $d_r(\mathbf{h}, \mathbf{r})$ of RotatE-RSCF can be expressed as:
\begin{align}
    \begin{split}
     d_r(\mathbf{h}, \mathbf{r}) &= \lVert \mathbf{h_r} \circ \mathbf{r_{ht}} - \mathbf{t_r} \rVert
    \end{split}
    \label{eq:RotatE-RSCF}
\end{align}
The score function $d_r(\mathbf{h}, \mathbf{r})$ of RESCAL-RSCF can be expressed as:
\begin{align}
    \begin{split}
     d_r(\mathbf{h}, \mathbf{r}) &= \lVert \mathbf{h_r} \mathbf{r_{h}} \mathbf{t} \rVert
    \end{split}
    \label{eq:RESCAL-RSCF}
\end{align}
In TDM, tail embeddings are not transformed according to the settings of SFBR in order to reduce computational costs. For the same reason, only the head entity is used for relation transformation.

\onecolumn
\begin{longtable}[t]{ll}
\hline
\small Relation Group & \small Relations \\ \hline

\multirow{8}{*}{\small position} & 
\fontsize{8.2pt}{10}\selectfont /sports/sports\_team/roster./basketball/basketball\_roster\_position/position \\
 & \fontsize{8.2pt}{10}\selectfont /soccer/football\_team/current\_roster./soccer/football\_roster\_position/position \\
 & \fontsize{8.2pt}{10}\selectfont /ice\_hockey/hockey\_team/current\_roster./sports/sports\_team\_roster/position \\
 & \fontsize{8.2pt}{10}\selectfont /sports/sports\_team/roster./american\_football/football\_historical\_roster\_position/position\_s \\
 & \fontsize{8.2pt}{10}\selectfont /sports/sports\_team/roster./baseball/baseball\_roster\_position/position \\
 & \fontsize{8.2pt}{10}\selectfont /sports/sports\_team/roster./american\_football/football\_roster\_position/position \\
 & \fontsize{8.2pt}{10}\selectfont /american\_football/football\_team/current\_roster./sports/sports\_team\_roster/position \\
 & \fontsize{8.2pt}{10}\selectfont/soccer/football\_team/current\_roster./sports/sports\_team\_roster/position \\ \hline
\multirow{14}{*}{\small currency} & \fontsize{8.2pt}{10}\selectfont/location/statistical\_region/gdp\_nominal\_per\_capita./measurement\_unit/dated\_money\_value/currency \\
 & \fontsize{8.2pt}{10}\selectfont/film/film/estimated\_budget./measurement\_unit/dated\_money\_value/currency \\
 & \fontsize{8.2pt}{10}\selectfont/business/business\_operation/operating\_income./measurement\_unit/dated\_money\_value/currency \\
 & \fontsize{8.2pt}{10}\selectfont/organization/endowed\_organization/endowment./measurement\_unit/dated\_money\_value/currency \\
 & \fontsize{8.2pt}{10}\selectfont/business/business\_operation/revenue./measurement\_unit/dated\_money\_value/currency \\
 & \fontsize{8.2pt}{10}\selectfont/business/business\_operation/assets./measurement\_unit/dated\_money\_value/currency \\
 & \fontsize{8.2pt}{10}\selectfont/location/statistical\_region/rent50\_2./measurement\_unit/dated\_money\_value/currency \\
 & \fontsize{8.2pt}{10}\selectfont/education/university/local\_tuition./measurement\_unit/dated\_money\_value/currency \\
 & \fontsize{8.2pt}{10}\selectfont/location/statistical\_region/gdp\_real./measurement\_unit/adjusted\_money\_value/adjustment\_currency \\
 & \fontsize{8.2pt}{10}\selectfont/education/university/domestic\_tuition./measurement\_unit/dated\_money\_value/currency \\
 & \fontsize{8.2pt}{10}\selectfont/education/university/international\_tuition./measurement\_unit/dated\_money\_value/currency \\
 & \fontsize{8.2pt}{10}\selectfont/location/statistical\_region/gdp\_nominal./measurement\_unit/dated\_money\_value/currency \\
 & \fontsize{8.5pt}{10}\selectfont/location/statistical\_region/gni\_per\_capita\_in\_ppp\_dollars./measurement\_unit/dated\_money\_value/currency \\
 & \fontsize{8.2pt}{10}\selectfont/base/schemastaging/person\_extra/net\_worth./measurement\_unit/dated\_money\_value/currency \\ \hline
\multirow{17}{*}{\small film production} & \fontsize{8.2pt}{10}\selectfont/film/film/costume\_design\_by \\
 & \fontsize{8.2pt}{10}\selectfont/film/film/executive\_produced\_by \\
 & \fontsize{8.2pt}{10}\selectfont/award/award\_winning\_work/awards\_won./award/award\_honor/award\_winner \\
 & \fontsize{8.2pt}{10}\selectfont/tv/tv\_program/program\_creator \\
 & \fontsize{8.2pt}{10}\selectfont/film/film/film\_art\_direction\_by \\
 & \fontsize{8.2pt}{10}\selectfont/film/film/music \\
 & \fontsize{8.2pt}{10}\selectfont/film/film/film\_production\_design\_by \\
 & \fontsize{8.2pt}{10}\selectfont/film/film/other\_crew./film/film\_crew\_gig/crewmember \\
 & \fontsize{8.2pt}{10}\selectfont/film/film/produced\_by \\
 & \fontsize{8.2pt}{10}\selectfont/tv/tv\_program/regular\_cast./tv/regular\_tv\_appearance/actor \\
 & \fontsize{8.2pt}{10}\selectfont/film/film/edited\_by \\
 & \fontsize{8.2pt}{10}\selectfont/film/film/written\_by \\
 & \fontsize{8.2pt}{10}\selectfont/film/film/personal\_appearances./film/personal\_film\_appearance/person \\
 & \fontsize{8.2pt}{10}\selectfont/film/film/story\_by \\
 & \fontsize{8.2pt}{10}\selectfont/film/film/cinematography \\
 & \fontsize{8.2pt}{10}\selectfont/film/film/dubbing\_performances./film/dubbing\_performance/actor \\
 & \fontsize{8.2pt}{10}\selectfont/film/film/production\_companies \\ \hline
\multirow{12}{*}{\small film actor} & \fontsize{8.2pt}{10}\selectfont/award/award\_nominee/award\_nominations./award/award\_nomination/nominated\_for \\
 & \fontsize{8.2pt}{10}\selectfont/tv/tv\_network/programs./tv/tv\_network\_duration/program \\
 & \fontsize{8.2pt}{10}\selectfont/film/special\_film\_performance\_type/film\_performance\_type./film/performance/film \\
 & \fontsize{8.2pt}{10}\selectfont/film/director/film \\
 & \fontsize{8.2pt}{10}\selectfont/tv/tv\_personality/tv\_regular\_appearances./tv/tv\_regular\_personal\_appearance/program \\
 & \fontsize{8.2pt}{10}\selectfont/film/film\_set\_designer/film\_sets\_designed \\
 & \fontsize{8.2pt}{10}\selectfont/tv/tv\_writer/tv\_programs./tv/tv\_program\_writer\_relationship/tv\_program \\
 & \fontsize{8.2pt}{10}\selectfont/film/actor/film./film/performance/film \\
 & \fontsize{8.2pt}{10}\selectfont/tv/tv\_producer/programs\_produced./tv/tv\_producer\_term/program \\
 & \fontsize{8.2pt}{10}\selectfont/media\_common/netflix\_genre/titles \\
 & \fontsize{8.2pt}{10}\selectfont/film/film\_distributor/films\_distributed./film/film\_film\_distributor\_relationship/film \\
 & \fontsize{8.2pt}{10}\selectfont/film/film\_subject/films \\ \hline
\multirow{8}{*}{\small people place} & \fontsize{8.2pt}{10}\selectfont/music/artist/origin \\
 & \fontsize{8.2pt}{10}\selectfont/people/person/places\_lived./people/place\_lived/location \\
 & \fontsize{8.2pt}{10}\selectfont/people/person/place\_of\_birth \\
 & \fontsize{8.2pt}{10}\selectfont/government/politician/government\_positions\_held./government/government\_position\_held/jurisdiction\_of\_office \\
 & \fontsize{8.2pt}{10}\selectfont/people/deceased\_person/place\_of\_death \\
 & \fontsize{8.2pt}{10}\selectfont/people/person/nationality \\
 & \fontsize{8.2pt}{10}\selectfont/people/deceased\_person/place\_of\_burial \\
 & \fontsize{8.2pt}{10}\selectfont/people/person/spouse\_s./people/marriage/location\_of\_ceremony \\ \hline
\multirow{8}{*}{\small film place} & \fontsize{8.2pt}{10}\selectfont/film/film/distributors./film/film\_film\_distributor\_relationship/region \\
 & \fontsize{8.2pt}{10}\selectfont/film/film/featured\_film\_locations \\
 & \fontsize{8.2pt}{10}\selectfont/film/film/release\_date\_s./film/film\_regional\_release\_date/film\_release\_region \\
 & \fontsize{8.2pt}{10}\selectfont/film/film/release\_date\_s./film/film\_regional\_release\_date/film\_regional\_debut\_venue \\
 & \fontsize{8.2pt}{10}\selectfont/film/film/country \\
 & \fontsize{8.2pt}{10}\selectfont/film/film/runtime./film/film\_cut/film\_release\_region \\
 & \fontsize{8.2pt}{10}\selectfont/tv/tv\_program/country\_of\_origin \\
 & \fontsize{8.2pt}{10}\selectfont/film/film/film\_festivals \\ \hline
\multirow{3}{*}{\small music role} & \fontsize{8.2pt}{10}\selectfont/music/group\_member/membership./music/group\_membership/role \\
 & \fontsize{8.2pt}{10}\selectfont/music/artist/track\_contributions./music/track\_contribution/role \\
 & \fontsize{8.2pt}{10}\selectfont/music/artist/contribution./music/recording\_contribution/performance\_role \\ \hline
\multirow{8}{*}{\small organization place} & \fontsize{8.2pt}{10}\selectfont/organization/organization/headquarters./location/mailing\_address/state\_province\_region \\
 & \fontsize{8.2pt}{10}\selectfont/organization/organization/place\_founded \\
 & \fontsize{8.2pt}{10}\selectfont/user/ktrueman/default\_domain/international\_organization/member\_states \\
 & \fontsize{8.2pt}{10}\selectfont/organization/organization/headquarters./location/mailing\_address/country \\
 & \fontsize{8.2pt}{10}\selectfont/people/marriage\_union\_type/unions\_of\_this\_type./people/marriage/location\_of\_ceremony \\
 &\fontsize{8.5pt}{10}\selectfont /base/schemastaging/organization\_extra/phone\_number./base/schemastaging/phone\_sandbox/service\_location \\
 & \fontsize{8.2pt}{10}\selectfont/government/legislative\_session/members./government/government\_position\_held/district\_represented \\
 & \fontsize{8.2pt}{10}\selectfont/organization/organization/headquarters./location/mailing\_address/citytown \\ \hline
\multirow{3}{*}{\small producer type} & \fontsize{8.2pt}{10}\selectfont/tv/tv\_producer/programs\_produced./tv/tv\_producer\_term/producer\_type \\
 & \fontsize{8.2pt}{10}\selectfont/film/film/other\_crew./film/film\_crew\_gig/film\_crew\_role \\
 & \fontsize{8.2pt}{10}\selectfont/tv/tv\_program/tv\_producer./tv/tv\_producer\_term/producer\_type \\ \hline
\multirow{5}{*}{\small award category} & \fontsize{8.2pt}{10}\selectfont/award/award\_category/winners./award/award\_honor/award\_winner \\
 & \fontsize{8.2pt}{10}\selectfont/award/award\_category/winners./award/award\_honor/ceremony \\
 & \fontsize{8.2pt}{10}\selectfont/award/award\_category/category\_of \\
 & \fontsize{8.2pt}{10}\selectfont/award/award\_category/nominees./award/award\_nomination/nominated\_for \\
 & \fontsize{8.2pt}{10}\selectfont/award/award\_category/disciplines\_or\_subjects \\ \hline
\caption{Clearly distinct relation groups that are selected from original TransE}
\label{tab:relation groups}\\
\end{longtable}
\twocolumn
\label{sec:appendix}

\end{document}